%% file: sample-manuscript.tex
\documentclass[manuscript,screen]{acmart}
\AtBeginDocument{%
  }

\setcopyright{acmlicensed}
\copyrightyear{2026}
\acmYear{2026}
\acmDOI{XXXXXXX.XXXXXXX}

\usepackage{booktabs}
\usepackage{multirow}
\usepackage{pifont} 
\newcommand{\cmark}{\ding{51}} \newcommand{\xmark}{\ding{55}}

\usepackage{colortbl}
\usepackage{xcolor}
\definecolor{colcol}{RGB}{173, 216, 230} % light blue, for example

\usepackage{nicematrix}
\usepackage{xcolor}
\definecolor{envcolor}{RGB}{173,216,230}

\newcommand{\todo}[1]{{\color{pink} TODO: #1}}

\newcommand{\massimo}[1]{{\color{red} Massimo: #1}}
\newcommand{\luca}[1]{{\color{purple} Luca: #1}}
\newcommand{\fontas}[1]{{\color{orange} Fontas: #1}}
\newcommand{\manuel}[1]{{\color{olive} Manuel: #1}}
\newcommand{\fabrizio}[1]{{\color{blue} Fabrizio: #1}}

\renewcommand{\todo}[1]{}
\renewcommand{\massimo}[1]{}
\renewcommand{\luca}[1]{}
\renewcommand{\fontas}[1]{}
\renewcommand{\manuel}[1]{}
\renewcommand{\fabrizio}[1]{}

\acmJournal{CSUR}

\begin{document}

%%\title{On-Device Learning in TinyML: A Survey Across Distribution Change Regimes}

\title{What changes after deployment? A survey on On-device Learning in TinyML}

\author{Massimo Pavan}
\authornote{This project has received funding from the European Union's Horizon Europe research and innovation programme under the HORIZON-JU-Chips-2024-1-IA grant agreement No 101194172. The project is also supported by its members France, Italy, Czech Republic, Germany, Sweden, Denmark, Greece, Spain, Netherlands, Portugal, Turkey, Switzerland. Views and opinions expressed are however those of the author(s) only and do not necessarily reflect those of the European Union or CHIPS JU. Neither the European Union nor the granting authority can be held responsible for them.}
\email{mapav@dtu.dk}
\orcid{0000-0002-5964-5685}
\affiliation{%
  \institution{Technical University of Denmark (DTU)}
  \city{Kongens Lyngby}
  \country{Denmark}
}

\author{Luca Pezzarossa}
\email{lpez@dtu.dk}
\orcid{0000-0002-0863-2526}
\affiliation{%
  \institution{Technical University of Denmark (DTU)}
  \city{Kongens Lyngby}
  \country{Denmark}
}

\author{Fabrizio Pittorino}
\email{fabrizio.pittorino@polimi.it}
\orcid{0000-0002-1919-6141}
\affiliation{%
 \institution{Politecnico di Milano}
 \city{Milan}
 \country{Italy}}

\author{Manuel Roveri}
\email{manuel.roveri@polimi.it}
\orcid{0000-0001-7828-7687}
\affiliation{%
 \institution{Politecnico di Milano}
 \city{Milan}
 \country{Italy}}

\author{Xenofon Fafoutis}
\email{xefa@dtu.dk}
\orcid{0000-0002-9871-0013}
\affiliation{%
  \institution{Technical University of Denmark (DTU)}
  \city{Kongens Lyngby}
  \country{Denmark}
}

\renewcommand{\shortauthors}{Pavan et al.}

\begin{abstract}
Machine learning models on microcontroller-class devices (TinyML) face a fundamental challenge: post-deployment distribution change undermines static models. On-device learning (ODL) addresses this by running the learning process directly on the device. The existing literature has not characterized how distribution change occurs or how different change types require different solutions.
Approximately 70 ODL works are surveyed under one principle: the distribution change regime. The survey analyzes how different types of distribution change influence the applications addressable on-device, the hardware employed, and the structure of the solutions. A persistent gap between methodological benchmarks and real-world deployment scenarios is also identified.
\end{abstract}




\received{--}
\received[revised]{--}
\received[accepted]{--}

\maketitle

\section{Introduction}

\input{Nomenclature}

The last decade has witnessed an unprecedented proliferation of intelligent embedded devices, from wearables monitoring vital signs to industrial sensors detecting early signs of mechanical failure. At the heart of this trend lies Tiny Machine Learning (TinyML): the discipline of deploying machine learning algorithms directly on highly constrained hardware, such as Microcontroller Units (MCUs), operating under tight budgets of memory, computation, and energy~\cite{warden_tinyml_2020}. By pushing ML inference to the extreme edge of the network (i.e., into the devices themselves, rather than delegating computation to remote servers) TinyML enables applications that are responsive, privacy-preserving, and functional even in the absence of network connectivity~\cite{banbury2021mlperf}.
Historically, the dominant paradigm in TinyML has been to separate the learning and inference phases entirely: models are trained on external hardware and subsequently deployed to the device, which is then responsible solely for running predictions. This separation keeps the computational burden on the device to a minimum, but it comes at a cost. Models trained offline on curated datasets frequently underperform once exposed to real-world conditions, as the data encountered after deployment rarely matches the distribution seen during training~\cite{widmer1996learning, lu2018learning}: in other words, the data distribution changes after deployment.
These shortcomings have motivated a growing body of work under the umbrella of \textit{On-Device Learning} (ODL)~\cite{rajapakse_intelligence_2023}, which seeks to execute not just inference but the learning process itself on the device, with the objective of adapting models to the distribution changes that occur after deployment.

Realizing this vision on TinyML hardware, however, requires confronting three constraints that together define ODL as a fundamentally distinct research challenge: memory and compute budgets must accommodate the learning process alongside inference, labeled data is scarce and costly to obtain on deployed devices, and solutions cannot be validated before they begin operating in the field. 
Despite these constraints, ODL substantially widens the scope of what TinyML systems can do: it enables fine-tuning to specific deployment conditions~\cite{hutchinson2017overcoming}, personalization to individual users or contexts~\cite{mcauley2022personalized}, and continuous adaptation to non-stationary data streams~\cite{de2021continual, ditzler2015learning}. The development of adaptive TinyML solutions is therefore an increasingly active research area, and one in need of careful analysis. However, despite the centrality of distribution change as the fundamental motivation for ODL, the existing literature has devoted surprisingly little effort to characterizing the different types of change that can occur after deployment, and prior surveys do not make explicit that different solutions operate under fundamentally different distribution change regimes. Motivated by this observation, we present a comprehensive survey of the Tiny ODL literature that, for the first time, places the nature of the distribution change at the center of its structure and analysis, using it as the primary lens through which solutions are characterized, compared, and evaluated.

Concretely, we distinguish three types of distribution change regimes. In the \textit{single-change} regime, the distribution shifts exactly once at deployment and then remains stable: the challenge is to adapt quickly and efficiently to the new conditions, so that effective inference can be performed for the rest of the device's life. In the \textit{concept drift} regime, the distribution shifts continuously and unpredictably throughout the operational life of the device, requiring the solution to adapt quickly to each new condition while discarding knowledge that is no longer relevant. In the \textit{continual learning} regime, new concepts are introduced incrementally over time, and the solution must accumulate new knowledge without forgetting what it has previously learned, a challenge known as catastrophic forgetting~\cite{de2021continual}. Together, these three regimes cover all the scenarios addressed in the ODL literature, and they serve as the primary organizing axis of this survey.

These three settings place fundamentally different demands on the applications that can be addressed and on the learning algorithm and the hardware we use to address them, and important patterns only emerge when works are compared within the same regime. For each regime, we analyze the literature across three complementary lenses: the \textit{application} contexts each regime naturally accommodates, the \textit{hardware} platforms suited to its computational demands, and the technical composition of the \textit{solutions} it requires. Beyond characterizing each regime individually, this structure exposes patterns that cut across all three regimes: first, which application domains tend to drive the use of particular hardware platforms, and second, which hardware constraints rule out entire families of learning algorithms, leaving only a narrow set of viable approaches for a given device.	

The rest of the paper is organized as follows. Section~\ref{sect:positioning} positions the survey with respect to prior work. Section~\ref{sect:prob} formalizes the ODL problem, defining the application requirements, device constraints, and solution components that constitute the three analytical lenses through which the literature is surveyed in Sections~\ref{sect:application}, \ref{sect:hardware}, and \ref{sect:technical} respectively. Section~\ref{sect:future} discusses open challenges and future directions, and Section~\ref{sect:conclusions} concludes the paper.

\input{tables/survey_comparison}

\begin{figure}
    \centering
    \includegraphics[width=0.5\linewidth]{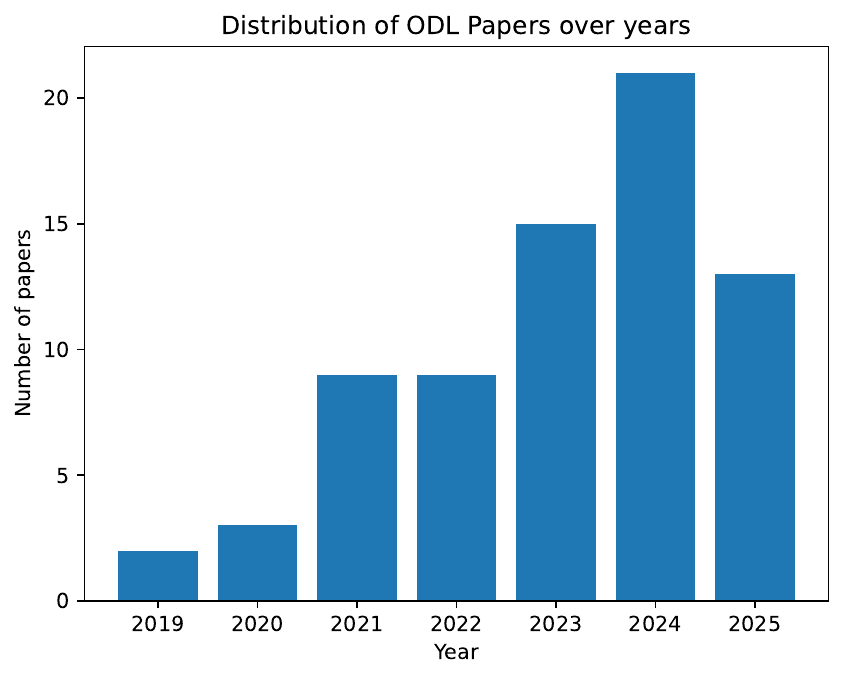}
    \Description{Bar chart showing the number of \textit{On-Device Learning} papers published per year in TinyML, from 2019 to 2025.}
    \caption{Number of papers published over the years in ODL for TinyML.}
    \label{fig:num_papers}
\end{figure}

\section{Scope and Positioning}
\label{sect:positioning}

Several surveys have recently addressed adjacent topics, but none covers the same combination of scope and perspective: prior works either target a broader hardware range than MCU-class devices or organize solutions along a technical axis that cuts across the distribution change regimes introduced here. Table~\ref{tab:survey_comparison} summarizes the key differences between related surveys and the present work.

Rajapakse et al.~\cite{rajapakse_intelligence_2023} is the closest in spirit, surveying both Over-The-Air (OTA) updates and ODL approaches for TinyML. Their taxonomy organizes ODL solutions along a technical axis, distinguishing offline learning from online learning approaches. While useful for characterizing algorithmic families, this axis cuts across the distribution change regime a solution operates under: a solution in which the distribution changes only once at the moment of deployment and one in which it can continuously change are grouped together, despite solving fundamentally different problems under fundamentally different constraints. Furthermore, the survey was published in 2023 and does not capture the substantial body of work published since. As visible in Figure~\ref{fig:num_papers}, most of the works present in the literature were published after 2023, highlighting the growing relevance of the field. Other more recent works are present in the literature, but they are less close in scope with respect to~\cite{rajapakse_intelligence_2023}. Abozaid et al.~\cite{abozaid_adaptive_2025} focus narrowly on the model-related aspects, offering detailed accounts of individual papers without a unifying taxonomic framework. Lourenço et al.~\cite{lourenco_-device_2025} restrict their scope to data stream learning for IoT, covering neural networks and decision trees as relevant model families, but without a focus on truly constrained hardware and without considering the hardware and application aspects.

The present survey differs from prior work in three respects. First, it is strictly scoped to MCU-class hardware and below, keeping hardware constraints central to the analysis. Second, with approximately 70 works surveyed, it covers a substantially larger portion of the literature than any prior effort. Third, and most distinctively, it organizes the entire literature around the distribution change regime, distinguishing between single-change, concept drift, and continual learning regimes, and analyzes each through three complementary lenses: application, hardware, and technical composition of the solutions.

\subsection{Literature Search and Inclusion Criteria}
\label{sect:inclusion}

The literature search was conducted using Google Scholar as the primary database, complemented by IEEE Xplore, ACM Digital Library, and arXiv. The search combined several keyword groups, including ``On-Device Learning", ``On-Device Training", ``Online Learning", ``Continual Learning", ``Transfer Learning", and ``Learning in the presence of concept drift", each paired with the terms ``tiny", ``MCU", or ``IoT" to restrict results to the TinyML domain. %Federated Learning was also included as a keyword combination, given its relevance as a complementary paradigm to ODL. 
The references of selected works were iteratively reviewed to identify additional relevant contributions not captured by the initial keyword search.

Works were included in the survey if they satisfied the following criteria:

\begin{itemize}
    \item The work targets hardware within the TinyML scope. Concretely, this corresponds to MCU-class devices (e.g., ARM Cortex-M series) or below, operating under tight RAM memory (in the order of the MB), frequency (<500 MHz), and power (<500 mW) budgets. Application-class processors (e.g., ARM Cortex-A series, as found in devices such as the Raspberry Pi) were considered outside this scope, unless the work explicitly frames its solution around the constraints of MCU-class deployment.
    \item The work addresses the learning phase of a machine learning algorithm, not inference only. \item The learning phase is executed or designed to be executed on the target device, not exclusively on external hardware.
\end{itemize}

\section{Problem Formulation for On-device Learning under distribution changes}
\label{sect:prob}

This section formalizes the on-device learning problem under distribution change, establishes the vocabulary and conceptual framework used throughout the survey, and defines the three distribution change regimes that categorize the surveyed works across the application, hardware, and solution lenses. A table summarizing the notation used in this section is provided in the appendix~\ref{app:notation}.

The goal of the ODL problem is to obtain a solution 
$\solution{}$\footnote{We adopt the term \textit{ODL solution} 
rather than \textit{ODL model}, as used in some prior work, to 
emphasize that a complete ODL solution must encompass not only 
the components for model learning and inference, but also those 
responsible for extracting and managing relevant features from 
the incoming data stream.} addressing an application, which 
comprises and enables the on-device inference and learning of a 
machine learning model over data drawn from a data-generating 
process $\dataprocess{}$ directly on a target device $\device{}$. The nature of $\dataprocess{}$, the requirements of the 
application, and the constraints of $\device{}$ together 
determine the space of viable solutions, and are 
formally characterized in the following subsections.

Section~\ref{sect:prob-dataproc} characterizes $\dataprocess{}$ 
and introduces the distribution change regime taxonomy that 
serves as the primary organizing principle of this survey: the 
nature of the change in $\dataprocess{}$ after deployment 
determines which regime a solution operates under. 
Section~\ref{sect:prob-app} formalizes the requirements that an application may impose on the solution and on the choice of $\device{}$. Section~\ref{sect:prob-device} characterizes $\device{}$ and the constraint it imposes on $\solution{}$. Section~\ref{sect:components} introduces the component
decomposition used throughout the technical analysis of the solutions,
together with the criteria for evaluating $\solution{}$ in each regime.

%============================= SECTIONS =================================

\subsection{The data generating process \texorpdfstring{$\dataprocess{}$}{P} 
and the distribution change regimes}
\label{sect:prob-dataproc}

Let $\dataprocess{}$ be a data generating process that, at each 
time instant $t$, provides a pair $(x^t, y^t)$ sampled from an 
unknown probability distribution $p^t(x, y)$, where $x$ is the 
data collected from an on-device sensor (e.g., an image or an 
audio clip) and $y$ its classification label. Without loss of
generality, the supervised information $y^t$ might not be
available at every time instant, reflecting the practical reality
of TinyML deployments where labeled data is scarce and learning
episodes are intermittent rather than continuous.

We assume that there exists a time instant $\deploytime{}$ when 
the proposed solution is deployed on an embedded device $\device{}$. 
We assume that before $\deploytime{}$, the unknown distribution is 
stationary: $\forall\, t < \deploytime{} \colon p^t(x, y) = 
p^{t+1}(x, y)$. This property might not hold for $t \geq 
\deploytime{}$, since the unknown distribution might change after 
deployment. It is worth noting that the change in $p^t(x, y)$ 
might affect the input $x$ (e.g., by the introduction of noise), 
the set of classes (e.g., class change), or the relationship 
between $x$ and $y$.
The set of realizations $\trainset = ((x^0, y^0), \ldots, 
(x^{\deploytime{}-1}, y^{\deploytime{}-1}))$ obtained from 
$\dataprocess{}$ before deployment constitutes a dataset that 
can be used for pre-training or initializing the various components of the 
solution $\solution{}$, while the set of realizations $\stream{} 
= ((x^{\deploytime{}}, y^{\deploytime{}}), \ldots)$ represents 
a stream of data arriving at the device after deployment, which 
can be used for the adaptation of the solution.

\subsubsection{Distribution change regimes}

Depending on whether and when a change in the distribution of $\dataprocess{}$ occurs during $\stream{}$, we can distinguish among three distribution change regimes, which together cover all the works present in the ODL literature:

\begin{itemize}
    \item \textit{Single-change regime:} The distribution of the data-generating process changes exactly once, at the moment of deployment $\deploytime{}$, i.e., $p^{\deploytime{}-1}(x, y) \neq p^{\deploytime{}}(x, y)$ and $\forall\, t > \deploytime{} \colon p^t(x, y) = p^{t+1}(x, y)$. Note that this regime also encompasses solutions trained entirely on-device without any pre-deployment data, i.e., solutions with $\trainset{} = \varnothing$. A typical example is a keyword spotting system that enrolls a user-defined wake word at deployment time and requires no further adaptation thereafter~\cite{rusci_self-learning_2025}.
    \item \textit{Concept drift regime:} The distribution of the data-generating process changes multiple times after deployment, i.e., $\exists\, t > \deploytime{} \colon p^{t-1}(x, y) \neq p^{t}(x, y)$. Changes in this regime are typically frequent and the focus is on designing solutions that adapt quickly to the new distribution, discarding or downweighting knowledge acquired under previous distributions. A typical example is a pressure sensor calibration model that continuously tracks measurement drift caused by environmental changes~\cite{saccani_-sensor_2024}.
    \item \textit{Continual learning regime:} This regime shares the same formal definition as the concept drift regime: $\exists\, t > \deploytime{} \colon p^{t-1}(x, y) \neq p^{t}(x, y)$. The distinction is not in the properties of $\dataprocess{}$, but in the assumptions made about the relevance of past knowledge and in the criteria used to evaluate solutions. Changes in this regime are typically less frequent and correspond to the introduction of genuinely new concepts or tasks. The focus is therefore on designing solutions that acquire new knowledge without forgetting previously learned information, a challenge commonly referred to as catastrophic forgetting~\cite{parisi2019continual}. A typical example is an EEG-based brain-machine interface that incrementally learns to recognize new motor commands while retaining the ones learned previously~\cite{mei_train--request_2024}.
\end{itemize}

While the concept drift and continual learning regimes share the same formal characterization of $\dataprocess{}$, treating them as a single category would group solutions designed under incompatible assumptions about the relevance of past knowledge, making direct comparison between them meaningless. The distinction is therefore maintained as a primary organizing principle throughout this survey.

The stationary case, in which no change occurs at any point in time, i.e., $\forall\, t > 0 \colon p^{t-1}(x, y) = p^{t}(x, y)$, is explicitly excluded from this survey. In such a setting, ODL provides no advantage over the train-then-deploy paradigm: any performance gain could be achieved more reliably by collecting additional labeled data and retraining offline. ODL is therefore only meaningful, and only surveyed here, in the presence of a change in $\dataprocess{}$ after $\deploytime{}$.

\subsection{The Application requirements}
\label{sect:prob-app}

An application is characterized by a data-generating process $\dataprocess{}$, by the task it addresses (e.g., classification, regression ...), and optionally by two technical requirements on the deployment: a power consumption requirement, i.e., the power drawn by $\device{}$ must remain within a budget imposed by the energy source and 
form factor of the deployment\footnote{Depending on the 
application, this may apply to average power over a defined 
operational window, to instantaneous peak power, or both.}, 
and an execution time requirement, i.e., each prediction must 
be produced within $\executionconstraint{}$ time units of the 
corresponding observation $\inp{}$ arriving at $\device{}$:
\[
\forall \ti > \deploytime{}  \quad \executiontime{}\!\left(\solution{}^{\ti{}}, \device{}\right) 
\leq \executionconstraint{},
\]

Where $\executiontime{}\!\left(\solution{}^{\ti{}}, \device{}\right)$ denotes the execution time of $\solution{}$ on $\device{}$ at time $\ti{}$. Satisfying these requirements involves two degrees of freedom: the choice of $\device{}$ and the design of $\solution{}$. The large majority of works in the literature fix $\device{}$ first and optimize the predictive performance of $\solution{}$ subject to the resulting constraints\footnote{The reverse is also 
possible: design $\solution{}$ first and identify the most 
efficient or cost-effective $\device{}$ capable of executing it. 
Co-optimization of both simultaneously is conceivable, but 
requires explicit weighting of cost, power consumption, and 
predictive performance against each other.}. The different applications addressed in the literature are analyzed in detail in Section~\ref{sect:application}.

\subsection{The Tiny Device \texorpdfstring{$\device{}$}{D}}
\label{sect:prob-device}

Let $\device{}$ be the target device on which $\solution{}$ is 
executed, chosen or designed to meet the application requirements 
introduced above. $\device{}$ is characterized by its RAM 
capacity $\memorydevice{}$, which bounds the amount of data, 
intermediate activations, and learnable parameters that can 
reside in memory simultaneously during execution, and by its 
clock frequency $\frequency{}$, which jointly determines 
$\executiontime{}$ alongside the algorithmic complexity of 
$\solution{}$. Once $\device{}$ is chosen, it imposes its own 
constraint on $\solution{}$ through $\memorydevice{}$:
\[
\memory{}_{\solution{}}^{\ti{}} \leq \memorydevice{},
\]
which must be satisfied at every time instant
$\ti{} \geq \deploytime{}$.
Note that $\memorydevice{}$ denotes the total on-chip RAM
of the device; the memory actually available to
$\solution{}$ is strictly smaller, as firmware, the
system stack, and runtime overhead consume a
non-negligible share of it. Nevertheless, most surveyed
works treat $\memorydevice{}$ as the effective memory
budget for the solution, and we follow this convention throughout the survey.
$\device{}$ is further characterized by a flash memory capacity, 
which determines the amount of static data, such as frozen model 
parameters and program code, that can be stored on the device. 
While flash capacity constrains the overall size of $\solution{}$, 
peak RAM is the binding feasibility constraint for ODL: unlike 
flash, which is written once at deployment, RAM must simultaneously 
accommodate the model parameters, the input data, the intermediate 
activations of the learning mechanism, and the data management 
buffer during the learning phase. Peak RAM is therefore the 
primary focus of the memory analysis throughout this survey.
The different devices employed in the literature are analyzed in detail in Section~\ref{sect:hardware}.

\subsection{The Components of an ODL Solution and its evaluation}
\label{sect:components}
\begin{figure}[ht]
    \centering
    \includegraphics[width=0.5\linewidth]{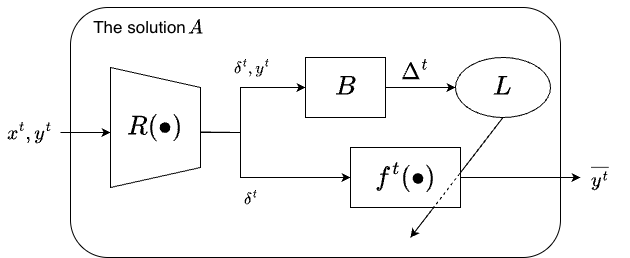}
    \Description{Block diagram of the general architecture of an ODL solution, showing the four components: dimensionality reduction block, data management block, ML algorithm, and learning mechanism, with arrows indicating their interactions.}
    \caption{The general architecture of an ODL solution
    $\solution{}$.}
    \label{fig:general}
\end{figure}

Any ODL solution in the surveyed literature can be characterized in terms of four components that communicate with each other as illustrated in Figure~\ref{fig:general}, regardless of the distribution change regime it is designed to operate in:

\begin{itemize}
    \item A \textit{dimensionality reduction block} 
    $\feblock{}(\bullet)$, which takes as input the raw data 
    $\inp{}$ and outputs a lower-dimensional representation 
    $\latentrep{}^t$, where $|\latentrep{}^t| \leq |\inp{}|$;
    \item A \textit{data management block} $\filterblock{}$, 
    which selects the instances to be passed to the learning 
    mechanism $\learningmech{}$. It is characterized by: (i) an 
    admission condition, determining whether a new instance 
    $\latentrep{}^t$ is stored in an internal buffer 
    $\filterbatch{}^t$ of maximum capacity $|\filterbatch{}|$; 
    (ii) the buffer $\filterbatch{}^t$ itself, which holds the 
    selected instances and evolves over time as new instances 
    are admitted; and (iii) a trigger condition, determining 
    when the learning mechanism $\learningmech{}$ is activated 
    on the entirety of $\filterbatch{}^t$;
    \item \textit{An ML algorithm} $\algo{}^t(\bullet)$, which 
    takes as input $\latentrep{}^t$ and outputs the prediction 
    $\pred{}$;
    \item \textit{A learning mechanism} $\learningmech{}$, which 
    uses the buffered instances $\filterbatch{}^t$ to update 
    $\algo{}^t$, with the objective of improving its performance 
    on future instances of $\stream{}$.
\end{itemize}

\subsubsection{Resource Requirements}

At each time instant $\ti{}$, each component has an associated 
memory requirement and an associated execution time, which 
together constitute the total memory requirement
$\memory{}_{\solution{}}^{\ti{}}$ and total execution time
$\executiontime{}\!\left(\solution{}^{\ti{}}, \device{}\right)$\footnote{The additive decomposition assumes sequential execution of the components, which holds for the large majority of the surveyed works. On parallel architectures, components may execute concurrently, in which case $\executiontime{}\!\left(\solution{}^{\ti{}}, \device{}\right)$ may need to be computed differently.} of $\solution{}$ at 
time $\ti{}$, the quantities that must satisfy the constraints 
introduced in Sections~\ref{sect:prob-app} 
and~\ref{sect:prob-device}:

\begin{equation}
\label{eq:memory}
\memory{}_{\solution{}}^{\ti{}} = \memory{}_{\feblock{}}^{\ti{}} + 
\memory{}_{\filterblock{}}^{\ti{}} + 
\memory{}_{\learningmech{}}^{\ti{}} + 
\memory{}_{\algo{}}^{\ti{}},
\end{equation}
\begin{equation}
\label{eq:exectime}
\executiontime{}\!\left(\solution{}^{\ti{}}, \device{}\right) =
\executiontime{}_{\feblock{}}^{\ti{}} +
\executiontime{}_{\filterblock{}}^{\ti{}} + 
\executiontime{}_{\learningmech{}}^{\ti{}} + 
\executiontime{}_{\algo{}}^{\ti{}}.
\end{equation}
Without loss of generality, at each time instant $\ti{}$, any 
component of $\solution{}$ can reduce to the identity function, 
making both its memory requirement and its execution time equal 
to zero. For instance, a solution with no dimensionality 
reduction block sets $\feblock{}(\bullet) = \text{id}(\bullet)$, 
yielding $\memory{}_{\feblock{}}^{\ti{}} = 
\executiontime{}_{\feblock{}}^{\ti{}} = 0$. Similarly, any block can be inactive at a given $\ti{}$, reducing its execution time to zero for that instant, while still maintaining a non-zero memory footprint. Finally, both 
$\memory{}_{\solution{}}^{\ti{}}$ and
$\executiontime{}\!\left(\solution{}^{\ti{}}, \device{}\right)$ may vary over time, 
depending on the architecture, parameters, and amount of data 
involved in the operations of each block.

With these edge cases (i.e., $\feblock{}(\bullet) = \text{id}(\bullet)$ and $|\filterbatch{}| = 1$) this decomposition covers all works surveyed in this paper. The boundary between $\feblock{}$ and $\algo{}^t$ is not always sharp in end-to-end trained networks, but the separation is present and meaningful in the large majority of surveyed works.

The different solutions components employed in the literature are analyzed in detail in Section~\ref{sect:technical}.

\subsubsection{Evaluation of Predictive Performance of ODL Solutions}
\label{sect:prob-eval}

Evaluating an ODL solution is fundamentally more complex than evaluating a static ML model. In the train-then-deploy paradigm, performance is measured once on a held-out test set after training is complete. In ODL, by contrast, the learning phase occurs during the operational life of $\device{}$, and performance evolves over time accordingly.

Since the model that will actually run on $\device{}$ does not yet exist at evaluation time, ODL works typically perform evaluation offline and off-device, using a \textit{held-out data stream} that simulates $\stream{}$, constructed from real data withheld for evaluation purposes. Let $\metric{}(\solution{}, \ti{})$ be a performance metric of $\solution{}$ at time $\ti{}$, such as classification accuracy or F1 score. The appropriate evaluation protocol depends on the distribution change regime.

In the single-change regime, $\solution{}$ is run on the held-out stream for a given number of steps, and final performance is measured on a separate held-out test set $\testset{}$ drawn from the same post-deployment distribution~\cite{pavan_tinysv_2025, rusci_-device_2023, ren_tinyol_2021}. This procedure can be repeated across multiple stream lengths to assess data efficiency.

In the concept drift regime, the held-out stream serves as both the source of learning updates and the basis for evaluation, following the prequential protocol~\cite{gama2013evaluating}, in which each sample is first used for inference and then for a learning update~\cite{disabato_tiny_2024, lourenco_dfdt_2025}. Performance is summarized as a running average of $\metric{}(\solution{}, \ti{})$ over the stream.

In the continual learning regime, a separate $\testset{}$ is maintained for each distribution encountered in $\stream{}$. Each $\testset{}$ is evaluated both immediately after $\solution{}$ finishes learning from the corresponding distribution, and again after exposure to all subsequent distributions~\cite{kwon_lifelearner_2023, ravaglia_tinyml_2021, mei_ultra-low_2025}. This longitudinal evaluation tracks knowledge acquisition and retention independently: a drop in performance on a previously evaluated $\testset{}$ indicates catastrophic forgetting~\cite{parisi2019continual}, while an improvement on a not-yet-seen $\testset{}$ indicates positive forward transfer.

\section{The Application Lens}
\label{sect:application}
% Short intro: method vs application as primary axis,
% regime as secondary axis within each part,
% forward pointer to complete pipeline section,
% note on the personalization vs environmental adaptation
% distinction deferred to summary
This section analyzes the application contexts in which ODL 
solutions have been deployed and evaluated. The literature 
is organized along two complementary axes. The primary 
distinction separates \textit{method-oriented} works, which 
propose algorithmic contributions validated on standard 
benchmark datasets without a specific application framing, 
from \textit{application-oriented} works, which are motivated 
by and evaluated on a specific real-world deployment scenario. 
Within each of these two categories, works are further 
organized by the distribution change regime in which they 
operate, following the same single-change, concept drift, and 
continual learning taxonomy established in 
Section~\ref{sect:prob-dataproc}.

\input{tables/benchmarks}

\subsection{Method-Oriented Works}
\label{sect:application-method}

Method-oriented works use standard benchmark datasets as a proxy for real-world scenarios; for this reason, Table~\ref{tab:benchmarks} lists those present in the surveyed literature, organized by regime and task.

\subsubsection{Single-Change Regime}
The majority of works evaluate on image classification 
benchmarks, including CIFAR-10, CIFAR-100, MNIST and 
their variants, and domain-specific collections such as 
Flowers, Food, Cars, CUB, and 
Pets~\cite{cai_tinytl_2021, lin_-device_2024, 
deutel_-device_2025, rub_tinyprop_2023, rub_advancing_2024, 
kwon_tinytrain_2024, rub_sparse_2025, pau_tinyrce_2023, 
pau_tinyrce_2023-1, pavan_tactile_2025, rub_drip_2025, 
disabato_incremental_2020, ren_tinyol_2021, sudharsan_edge2train_2020, 
sudharsan_ml-mcu_2022, sudharsan_train_2021}. A number 
of these works additionally include audio benchmarks, 
in particular the Google Speech Commands 
dataset~\cite{rub_tinyprop_2023, rub_advancing_2024, 
rub_sparse_2025, pau_tinyrce_2023, rub_drip_2025, 
ren_-device_2024} and DCASE2020~\cite{rub_tinyprop_2023, 
rub_advancing_2024}, as well as Inertial Measurement Units (IMUs) sensor 
benchmarks for Human Activity Recognition (HAR) such as 
PAMAP2, SHL, and CWRU~\cite{pau_tinyrce_2023-1} and 
UCI-HAR~\cite{rub_advancing_2024, rub_sparse_2025}, 
reflecting the relevance of these tasks in the TinyML 
community. Finally, a smaller group of works evaluates 
on tabular datasets such as Iris, heart disease, and 
cancer~\cite{sudharsan_edge2train_2020, 
sudharsan_ml-mcu_2022, sudharsan_train_2021}. While 
these datasets are simpler than image or audio benchmarks, 
they are representative of the sensor data commonly 
encountered in IoT applications, making them a relevant, 
even if underused, evaluation domain for TinyML.

\subsubsection{Concept Drift Regime}
The method-oriented works in the concept drift regime 
are fewer than in the single-change setting.
\cite{disabato_tiny_2024}, \cite{ma_-demand_2025}, 
and~\cite{pavan_tybox_2024} evaluate on image and audio 
benchmarks from the broader machine learning community, 
including Google Speech Commands (GSC), MNIST, CIFAR-10 and ImageNet, where concept drift is simulated through synthetic corruptions of standard datasets (e.g., CIFAR-10-C, ImageNet-C, 
SHIFT). In contrast,~\cite{lourenco_dfdt_2025} evaluates 
exclusively on classic stream learning benchmarks from 
the data stream mining community, including NOAA, ELEC, 
RIALTO, POSTURE, COVER, and POKER, which feature 
naturally occurring distribution shifts rather than
synthetic ones, at the cost of addressing tasks that are less representative of typical TinyML deployment scenarios.

\subsubsection{Continual Learning Regime}
The method-oriented works in the continual learning 
regime evaluate primarily on image classification 
benchmarks, with CORe50~\cite{ravaglia_tinyml_2021, 
rub_continual_2024, vorabbi_enabling_2024}, CIFAR-10 
and CIFAR-100~\cite{kwon_lifelearner_2023, 
rub_continual_2024, vorabbi_enabling_2024, 
wibowo_12_2024}, and 
MiniImageNet~\cite{kwon_lifelearner_2023} being the 
most commonly used. \cite{rub_continual_2024} also 
includes MNIST as an additional benchmark. CORe50 is 
particularly well suited to continual learning 
evaluation as it was specifically designed for object 
recognition under continuous learning conditions, making 
it more representative of real deployment scenarios than 
generic image classification benchmarks. Beyond image 
classification,~\cite{kwon_lifelearner_2023} 
and~\cite{rub_continual_2024} additionally evaluate on appropriately modified versions of
GSC and UCI-HAR respectively, providing some evidence 
of generalization beyond the image domain.

\subsection{Application-Oriented Works}
\label{sect:application-real}

Application-oriented works are motivated by and evaluated on specific real-world deployment scenarios, making the application domain the primary axis of comparison; Table~\ref{tab:applications} lists those present in the surveyed literature, organized by regime and task.

\input{tables/applications}

\subsubsection{Single-Change Regime}

\paragraph{Keyword Spotting.}
Keyword Spotting (KWS) is the most represented application 
in the single-change regime, addressed from three 
complementary angles. Personalization to a specific 
user's speech characteristics is addressed 
by~\cite{cioflan_boosting_2024}, which adapts a 
pre-trained model using a small number of labeled 
utterances from the target user. Noise adaptation to 
the acoustic conditions of the deployment context 
is addressed by~\cite{cioflan_towards_2022, 
cioflan_-device_2024}, which fine-tune a subset of 
model parameters using recordings collected on-site. 
The enrollment of new, user-defined keywords is 
addressed by~\cite{rusci_-device_2023, 
rusci_few-shot_2023, rusci_self-learning_2025, 
chauhan_exploring_2022}, where the challenge is to 
recognize previously unseen words from a handful of 
recordings: some works rely on few-shot supervised 
learning~\cite{rusci_-device_2023, rusci_few-shot_2023, 
chauhan_exploring_2022}, while others eliminate the 
labeling requirement through self-supervised 
pseudo-labeling of the incoming audio 
stream~\cite{rusci_self-learning_2025}. 
\cite{pavan_tinysv_2025} extend the KWS personalization 
setting to speaker verification, combining keyword 
detection with user identity authentication in a 
two-stage on-device pipeline.

\paragraph{Human Pose Estimation.}
\cite{cereda_-device_2024, cereda_training_2024} 
demonstrate on-device self-supervised fine-tuning for 
human pose estimation aboard nano-drones, where the 
model adapts to the visual characteristics of a new 
deployment context. The supervision signal is derived 
from a secondary on-board sensor available only during 
the adaptation phase, making the approach applicable 
without any human annotation.

\paragraph{Human Activity Recognition.}
\cite{craighero_-device_2024} address personalization 
of IMU-based HAR on STM32 microcontrollers, 
demonstrating that fine-tuning with a small 
user-specific labeled dataset significantly reduces 
inter-subject variability. Both a publicly available 
benchmark and a proprietary collected dataset are used 
for evaluation, providing evidence of real-world 
applicability beyond standard benchmarks. 

\paragraph{Biosignal Processing.}
Physiological signals are inherently person-specific, 
making them natural candidates for on-device 
personalization. For ECG anomaly 
detection,~\cite{abdennadher_fixed_2021} train a 
personalized model from scratch on the MIT-BIH 
arrhythmia dataset, a clinically validated benchmark 
that closely reflects real deployment conditions. For 
sEMG-based applications,~\cite{burrello_tackling_2021} 
and~\cite{benatti_online_2019} address gesture 
recognition personalization, where cross-session 
variability due to electrode shift is the primary 
challenge, while~\cite{zanghieri_semg-driven_2024} 
extend this to hand dynamics estimation on the HYSER 
dataset, demonstrating that incremental online training 
achieves cross-day accuracy comparable to offline 
training.

\paragraph{Anomaly Detection.}
Anomaly detection appears across multiple deployment 
contexts in the single-change regime, sharing the 
common characteristic that normality must be learned 
on-device since it cannot be defined a priori. 
\cite{pereira_-device_2023} address industrial time 
series anomaly detection by fitting an analytical 
model directly from unlabeled post-deployment data. 
\cite{ren_tinyol_2021} demonstrate on-device anomaly 
detection for a fan monitoring application, adapting 
an autoencoder to the normal operating conditions of 
the deployment platform from collected data. 
\cite{silva_adaptive_2023} address driver behavior 
anomaly detection, where normality is user-dependent 
and learned through unsupervised on-device clustering 
of vehicle sensor data.

\paragraph{Presence and Occupancy Detection.}
\cite{renzi_online_2025} address occupancy detection 
from ambient environmental sensors in smart buildings, 
providing a systematic comparison of online learning 
algorithms on resource-constrained devices for this 
task.

\paragraph{Gesture Classification.}
\cite{chowdhary_-sensor_2023} tackle on-sensor gesture 
classification in an extremely constrained setting, 
learning a personalized model from scratch within 8~kB 
of memory from proprietary collected data.
    
\subsubsection{Concept Drift Regime}
\paragraph{Sensor Calibration.}
Sensor calibration is the only real-world application 
addressed in the concept drift regime. Physical 
sensors exhibit continuous drift over time due to 
aging, temperature variations, and mechanical stress, 
making calibration a genuinely non-stationary problem 
that requires the learning mechanism to remain active 
throughout the operational life of the device.
\cite{pau_learning_2025} address IMU calibration 
under thermal stress, learning a compensation model 
on the sensor's integrated processor from data 
collected under controlled temperature variations, 
with additional validation on the EuRoC MAV dataset 
under dynamic conditions. \cite{saccani_-sensor_2024} 
address pressure sensor calibration, learning a 
compensation model from sporadic reference 
measurements available during deployment, without 
requiring any offline recalibration procedure.

\subsubsection{Continual Learning Regime}
\label{sect:application-real-cl}
\paragraph{Brain-Machine Interfaces.}
EEG-based Brain-Machine Interfaces (BMI) are a natural 
application for continual learning personalization, 
as neural signals exhibit strong inter-session 
variability that causes performance degradation over 
time, and the set of mental states or motor commands 
the system must recognize may expand as the user's 
needs evolve. \cite{mei_train--request_2024} and 
\cite{mei_ultra-low_2025} address this by enabling 
continual on-device adaptation of EEG classification 
models to the evolving signal characteristics of a 
specific user, demonstrating that catastrophic 
forgetting can be mitigated while maintaining the 
ultra-low power requirements of wearable BMI devices. 
Both works evaluate on a dataset collected from 
wearable EEG hardware, with~\cite{mei_ultra-low_2025} 
additionally validating on a novel collected dataset.

\paragraph{Human Activity Recognition.}
Class-incremental learning for HAR addresses the 
scenario where new activities must be added to the 
model after deployment without retraining from 
scratch. \cite{leite_resource-efficient_2022} address 
class-incremental HAR from IMU data, evaluating on 
four publicly available benchmarks including 
Opportunity, PAMAP2, Skoda, and MHEALTH, and 
demonstrating resource-efficient expansion of the 
model as new activity classes are introduced.

\subsection{Findings}
\label{sect:application-summary}

\subsubsection{Method-Oriented vs Application-Oriented Works}
A striking observation that emerges from the comparison 
between method-oriented and application-oriented works 
is the limited overlap in the tasks and datasets they 
address. KWS and, to a lesser extent, IMU-based HAR are the only tasks that appear consistently in both categories, with GSC and UCI-HAR serving as their respective shared benchmarks. 

This misalignment is more pronounced 
in TinyML than in conventional machine learning, and 
reflects two concurrent structural factors. On one 
side, the application domains that are most studied 
in the methodological literature, primarily image 
classification, are too memory and computationally 
demanding to build relevant real-world TinyML 
applications with current hardware. On the other 
side, many of the real-world application domains 
that do fit within TinyML constraints, such as 
biosignal processing, sensor calibration, and 
anomaly detection, lack standardized benchmarks 
that methodological works can adopt for fair 
comparison. The result is a field where algorithmic 
progress and application development are advancing 
largely in parallel, with limited cross-fertilization.

\subsubsection{Personalization, Environmental Adaptation, and Sensor Calibration}
The application-oriented works in this survey address 
three broad categories of real-world challenges. 
Personalization tasks, the most represented, aim to 
adapt a model to a specific individual, spanning 
domains such as KWS,  
HAR, biosignal processing, and driver behavior 
analysis. Environmental adaptation tasks address 
distribution shifts originating from the deployment 
environment rather than from the user, as in the case 
of acoustic noise adaptation and visual perception 
aboard drones. Sensor calibration emerges as a 
distinct category, particularly in the concept drift 
regime, where the continuously drifting output of 
physical sensors must be compensated on-device 
throughout the operational lifetime of the device.

\subsubsection{The Absence of Explicit Application Requirements}
Despite the centrality of execution time and power 
consumption as application-level constraints, as 
established in Section~\ref{sect:prob-app}, the 
application-oriented works surveyed here rarely state 
the requirements they are targeting explicitly. In industrial
deployments, power budgets and latency constraints
($\executionconstraint{}$) are strict and non-negotiable. In research works, 
they are typically left implicit, with the choice of 
hardware class serving as a proxy.
This implicit convention has a structural consequence: 
without benchmarks that fix an explicit power budget 
and latency constraint $\executionconstraint{}$, the 
co-optimization of $\device{}$ and $\solution{}$ 
becomes practically infeasible, and hardware and 
algorithmic choices are made sequentially rather than 
jointly. Establishing benchmarks that specify explicit 
deployment requirements alongside task and dataset 
would be a concrete step toward enabling this 
co-optimization, and is discussed further in 
Section~\ref{sect:future}.

\section{The Hardware Lens}
\label{sect:hardware}

This section analyzes the hardware platforms on which 
the surveyed ODL solutions have been deployed and 
evaluated. As established in Section~\ref{sect:prob-device}, 
the feasibility of any ODL solution is fundamentally 
constrained by the on-device memory budget $\memorydevice{}$ and the clock frequency $\frequency{}$. The hardware choices made across the 
literature therefore reveal both the current boundaries 
of what ODL can achieve on real devices and the gap 
between algorithmic contributions evaluated on capable
platforms and solutions deployable on the most
constrained hardware.

The hardware platforms reported across the surveyed 
works can be grouped into three classes based on their 
computational architecture: \textit{ultra-constrained processors} 
(Section~\ref{sect:hardware-uc}), \textit{standard MCUs} 
(Section~\ref{sect:hardware-mcu}), and \textit{PULP platforms} 
(Section~\ref{sect:hardware-pulp}). These classes are 
not merely points on a performance spectrum: each 
represents a qualitatively different architectural 
design philosophy, with distinct implications for which ODL solutions are feasible. A non-negligible fraction 
of the surveyed works do not evaluate on a specific 
hardware platform, and a smaller number target platforms 
outside the TinyML scope defined in 
Section~\ref{sect:prob-device}, such as Cortex-A 
processors, Graphics Processing Units (GPUs), and smartphones. These works are 
excluded from the analysis carried out in this section but are included in the application and solution analyses, where their algorithmic contributions remain relevant.

\input{tables/hardware_ultra_agg}

\subsection{Ultra-Constrained Processors}
\label{sect:hardware-uc}

Ultra-constrained processors include platforms lacking 
a hardware floating-point unit, such as Cortex-M0~\cite{arm_cortex_m0_trm} based 
MCUs, with $\frequency{} \leq 100$~MHz, and the In-Sensor Processing Unit (ISPU - STMicroelectronics LSM6DSO16IS~\cite{st_lsm6dso16is_datasheet}). The ISPU integrates 
a hardware Floating-Point Unit (FPU) but operates at only $\frequency{} = 5$--$10$~MHz, placing 
it in the same practical regime as FPU-less devices. 
It is also architecturally unique in that it is embedded 
directly inside the sensor package, eliminating data 
movement between sensing and processing entirely. The 
constraints we just described, combined with a very small $\memorydevice{}$, imposes the tightest memory and computational
budgets of the three classes and constrains the 
algorithmic choices available, as discussed in 
Section~\ref{sect:technical}. 

Devices in this class span $\frequency{} \in [5~\text{MHz},\, 133~\text{MHz}]$ (ISPU to RP2040), with
$\memorydevice{}$ ranging from 8~kB to 264~kB,
as summarized in Table~\ref{tab:hardware_ultra_agg}.
Active power consumption is rarely reported in this 
class. Based on datasheet figures, the ISPU draws 
approximately 1~mW in active mode, while Cortex-M0 
based boards typically fall in the 10--50~mW range, 
depending on the specific board and supply voltage.

Cortex-M0 based boards appear in both method-oriented 
and application-oriented works. Their widespread 
availability and low cost make them a natural evaluation 
target for method contributions seeking to demonstrate 
feasibility on constrained hardware, as well as for 
application-driven deployments where the algorithmic 
and power budget is tight. The ISPU, by contrast, 
appears exclusively in application-oriented works, 
consistent with its nature as a purpose-built in-sensor 
processor: its architectural constraints make it 
unsuitable as a benchmark device for algorithmic 
contributions, which typically require more flexibility 
in terms of model architecture and evaluation tooling.

The surveyed works on ultra-constrained hardware are 
almost entirely concentrated in the single-change 
regime, with a single exception in concept 
drift~\cite{pau_learning_2025} and no works in the 
continual learning regime. This concentration reflects 
the constraints of the class: the algorithmic approaches 
feasible on these platforms can handle a single bounded 
adaptation phase, but sustaining a continuously active 
learning mechanism throughout the operational life of 
$\device{}$ would require memory and computational 
resources beyond what they offer.

\subsection{Standard MCUs}
\label{sect:hardware-mcu}

\input{tables/hardware_mcu}

Standard MCUs include general-purpose single-core 
processors equipped with a hardware FPU, such as 
Cortex-M4~\cite{arm_cortex_m4_trm} and Cortex-M7~\cite{arm_cortex_m7_trm} based platforms, with $\frequency{} \in [80,\, 480]$~MHz. These devices follow 
a conventional scalar pipeline architecture with no 
hardware-level optimization for neural network 
workloads, and represent the workhorse class of 
TinyML, supported by a mature software ecosystem and 
available across a wide range of off-the-shelf boards. 

The presence of an FPU and a larger $\memorydevice{}$,
typically between 256~kB and a few MB, makes
backpropagation-based learning feasible on this class, 
provided that appropriate memory optimizations are 
applied as discussed in Section~\ref{sect:technical}. 
Devices in this class span $\frequency{} \in [64,\, 480]$~MHz, with $\memorydevice{} \in [60~\text{kB},\, 512~\text{kB}]$ across the surveyed works.
Active power figures are not directly reported in any of the surveyed works in this class; all values in Table~\ref{tab:hardware_mcu} are estimated from datasheet active-mode figures. Based on these estimates, Cortex-M4 boards typically draw 20--60~mW, while Cortex-M7 boards fall in the 100--600~mW range, reflecting genuine differences in $\frequency{}$ and supply voltage across device families rather than workload variation, as active power is a property of the silicon.

Standard MCUs are the most balanced class in terms 
of regime coverage, with works present in all three 
regimes. The single-change regime is by far the most 
represented, covering both application-oriented works 
and methodological contributions exploring 
backpropagation optimization~\cite{lin_-device_2024, 
khan_dacapo_2023, patil_poet_2022} and forward-only 
learning~\cite{pau_tinyrce_2023, pau_tinyrce_2023-1, 
disabato_incremental_2020, ren_-device_2024}. Within 
this regime, contributions are distributed fairly 
evenly across Cortex-M4 and Cortex-M7 cores, and 
between method-oriented and application-oriented works, 
with no clear concentration in either dimension. In the concept drift regime, standard MCUs are the most represented hardware class~\cite{disabato_tiny_2024, pavan_tybox_2024}: concept drift solutions rely on lightweight incremental algorithms that neither require replay buffers nor incur the backpropagation overhead of continual learning, making standard MCUs a natural fit. In the continual learning regime, by contrast, only one work targets a standard MCU~\cite{kwon_lifelearner_2023}, with most solutions pushed toward more powerful devices by the memory demands of maintaining and replaying past distributions.

\subsection{PULP Platforms}
\label{sect:hardware-pulp}

\input{tables/hardware_pulp}

PULP platforms include multi-core parallel ultra-low-power 
architectures such as the commercial platforms GAP8 and 
GAP9, and the research prototypes Mr.~Wolf and VEGA. 
What distinguishes this class is its parallel multi-core architecture: instead of a single general-purpose core, PULP platforms feature a cluster of up to nine small cores operating in parallel over a shared memory, allowing them to process the same workload faster and at a lower energy cost per operation than a single high-performance core. 

Individual cores span $\frequency{} \in [100, 400]$~MHz
and $\memorydevice{} \in [15~\text{kB},\, 64~\text{MB}]$
across the surveyed works, reflecting the wide range
of deployment scenarios this class supports,
as summarized in Table~\ref{tab:hardware_pulp}. 
Active power is more consistently reported in this 
class than in the others. GAP9 figures range from 
19 to 66~mW during single-change 
workloads~\cite{cereda_training_2024, cereda_-device_2024, 
cioflan_-device_2024, rusci_-device_2023} and from 
21 to 50~mW for continual learning 
applications~\cite{mei_train--request_2024, 
mei_ultra-low_2025}, with \cite{wibowo_12_2024} 
reporting 45~mW. Mr.~Wolf draws approximately 
10.4~mW~\cite{benatti_online_2019}. VEGA and GAP8 
works do not report active power figures.

In the single-change regime, PULP is heavily used for 
audio and biosignal applications, with GAP9 being the 
dominant platform~\cite{cereda_training_2024, 
cereda_-device_2024, cioflan_-device_2024, 
cioflan_towards_2022, rusci_self-learning_2025, 
rusci_-device_2023, zanghieri_semg-driven_2024, 
rusci_self-incremental_2025}, alongside GAP8 for an 
earlier sEMG work~\cite{burrello_tackling_2021}, VEGA 
for KWS personalization~\cite{cioflan_boosting_2024}, 
and Mr.~Wolf for EMG gesture 
recognition~\cite{benatti_online_2019}.
In the concept drift regime, no PULP works were 
identified. Unlike the absence of ultra-constrained 
devices in continual learning, which can be explained 
by hardware limitations, the absence of PULP in 
concept drift is more likely a reflection of the 
different research communities that have historically 
addressed these two problems, rather than a fundamental 
feasibility constraint.
The continual learning regime is where PULP platforms 
are most distinctive: they account for the large 
majority of continual learning works with reported 
hardware~\cite{mei_train--request_2024, mei_ultra-low_2025, 
wibowo_12_2024, ravaglia_tinyml_2021}, reflecting the 
substantially higher memory and computational 
requirements of continual learning algorithms compared 
to single-change and concept drift solutions.

\subsection{Findings}
\label{sect:hardware-summary}

\subsubsection{Hardware Capability and Regime Complexity}
A clear correlation emerges between hardware capability 
and regime complexity. Ultra-constrained processors 
appear almost exclusively in the single-change regime, 
where the learning phase is bounded and lightweight 
algorithms suffice. Standard MCUs span all three 
regimes but are thinly represented in continual 
learning, where their $\memorydevice{}$ is often
insufficient to sustain an active learning mechanism 
over the device lifetime. PULP platforms dominate 
the continual learning regime, being the only class 
capable of supporting its most demanding solutions. 
This monotonic relationship suggests that the choice 
of hardware is not independent of the target regime: 
more complex regimes impose stricter requirements, 
and pushing ODL further into non-stationary regimes 
will likely require hardware capabilities beyond 
what standard MCUs can currently offer.

A complementary pattern emerges along the 
method-application axis. General-purpose ARM-based 
boards, whether Cortex-M0, M4, or M7, are the 
preferred platform for methodological contributions, 
favored for their wide availability, low cost, and 
reproducibility. More specialized platforms, namely 
the ISPU and PULP devices, appear predominantly in 
application-oriented works, where the hardware 
properties are themselves part of the contribution. 
This divide between algorithmic research, which uses 
commodity hardware as a neutral substrate, and 
system-level research, which develops the solution in tight coupling with the target platform, is a structural feature of the current ODL literature.

\subsubsection{Hardware Reporting in the Literature}
Comparing hardware platforms in a solution-agnostic way requires metrics that depend solely on $\device{}$, independently of the algorithmic choices made on top of it. The metrics most commonly reported in the surveyed works, namely peak RAM usage $\memorydevice{}$, energy per learning event, and latency $\executiontime{}\!\left(\solution{}^{\ti{}}, \device{}\right)$, all depend jointly on $\device{}$ and $\solution{}$, as established in Section~\ref{sect:components}, and are therefore unsuitable for this purpose. These metrics are instead examined in depth in Section~\ref{sect:technical-summary}.

In this chapter, we instead characterize each platform by three solution-agnostic metrics: RAM capacity $\memorydevice{}$, clock frequency $\frequency{}$, and peak instantaneous active power consumption, all of which are properties of the silicon rather than of the workload. These metrics describe what a device offers, but not how efficiently it executes a given ODL solution. 

This points toward a broader gap: the absence of fixed-solution benchmarks that, by holding $\solution{}$ constant and varying $\device{}$, would make energy consumption, latency, and peak RAM usage comparable across platforms and enable principled hardware selection. This is discussed further as an open challenge in Section~\ref{sect:future}. 

\subsubsection{PULP Platforms and the Hardware Landscape for ODL}
The dominance of PULP platforms in the continual 
learning regime deserves closer examination, 
particularly in light of a class of devices that 
sits between standard MCUs and PULP but has not 
appeared in the surveyed literature: MCUs equipped 
with a dedicated neural network accelerator, such 
as the ARM Cortex-M55 paired with an Ethos-U55~\cite{arm_ethos_u55_pb} or 
Ethos-U65~\cite{arm_ethos_u65_pb} Neural Processing Unit (NPU). These platforms retain the 
single-core scalar architecture of a standard MCU 
but add a fixed-function hardware block accelerating 
specific neural network operations, and could in principle offer a middle ground between standard MCUs and PULP platforms. Their absence from 
the ODL literature is likely explained by the fact 
that their software ecosystems have been developed 
almost exclusively for inference, with no support 
for gradient computation or on-device training, and 
that they were not widely available at the time of 
the earliest works in this survey.

PULP platforms have instead accumulated concrete 
advantages explaining their prominence. A dedicated 
ODL research community has formed around GAP9 and 
VEGA, producing shared methodology and open-source 
tools such as DORY~\cite{burrello2020dory} and PULP-TrainLib~\cite{Nadalini2022PULPTrainLib} that directly 
expose training primitives. PULP also offers a 
contained power envelope with substantially higher 
computational throughput than standard MCUs, and 
its flexible architecture, where active cores, 
memory banks, and off-chip memory can be configured 
independently, makes it adaptable across a wide 
range of deployment scenarios.

These advantages are accompanied by significant 
practical limitations. PULP platforms are 
considerably more complex to program than standard 
MCUs, requiring familiarity with cluster execution 
models, direct memory access management, and multi-bank memory 
hierarchies. More fundamentally, no commercial 
off-the-shelf development kit is currently 
available, restricting access to research groups 
with direct connections to the platforms' 
developers, and their unit cost remains 
substantially higher than commodity MCU solutions.

\section{The Solution Lens}
\label{sect:technical}

This section analyzes the technical composition of 
the ODL solutions surveyed in this work, focusing 
on how the components described in 
Section~\ref{sect:components} are implemented in 
practice. As argued in Section~\ref{sect:prob-dataproc}, 
solutions operating under different distribution 
change regimes are not directly comparable, and the 
analysis is organized accordingly. The primary axis 
is the learning paradigm: Section~\ref{sect:technical-bl} 
covers \textit{batch learning solutions}, in which 
$\learningmech{}$ executes exactly once on a 
collected dataset before $\solution{}$ transitions 
permanently to inference, and which by construction 
are limited to the single-change regime. 
Section~\ref{sect:technical-il} covers \textit{incremental 
learning solutions}, in which $\learningmech{}$ 
remains active throughout the operational life of 
$\device{}$. Within Section~\ref{sect:technical-il}, 
the secondary axis is the distribution change 
regime, following the single-change, concept drift, 
and continual learning taxonomy of 
Section~\ref{sect:prob-dataproc}.

\subsection{Batch Learning Solutions}
\label{sect:technical-bl}

\begin{figure}[hbtp]
    \centering
    \includegraphics[width=0.5\linewidth]{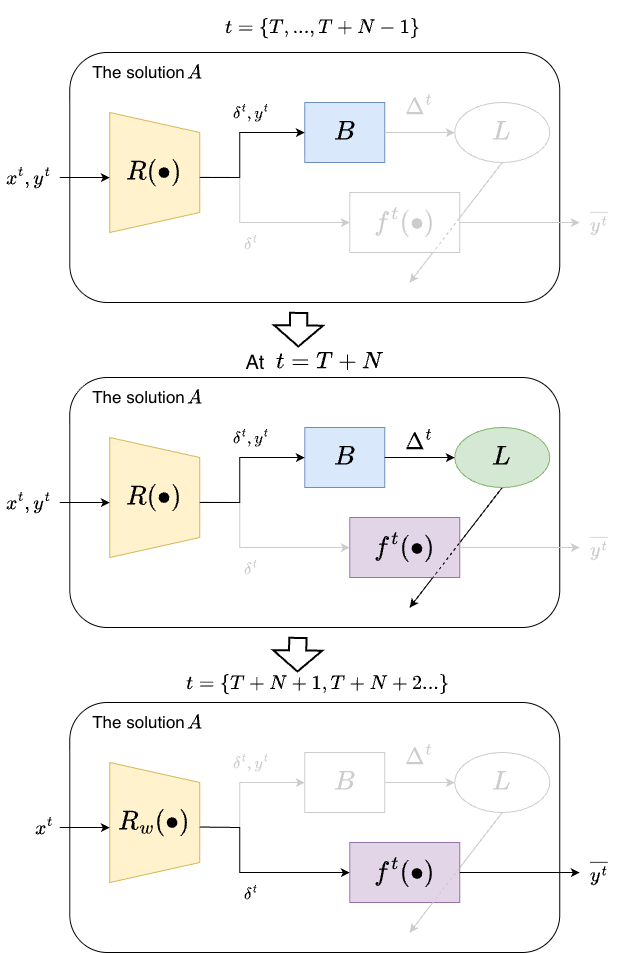}
    \Description{Diagram showing the temporal sequence of active components in a batch learning ODL solution: the data management block collects data, then the learning mechanism executes once, after which only the dimensionality reduction block and ML algorithm remain active for inference.}
    \caption{The sequence of configurations of a
    batch learning solution $\solution{}$. Colors indicate that the component is active at time $t$.}
    \label{fig:finite_learning}
\end{figure}

In batch learning solutions, as illustrated in Figure \ref{fig:finite_learning}, $\filterblock{}$ first 
collects a set of $\adaptsetdim{}$ instances from 
$\stream{}$ into the buffer $\filterbatch{}^t$. Once 
the trigger condition fires, $\learningmech{}$ is 
executed exactly once on $\filterbatch{}^t$, producing 
a definitive version of $\algo{}^t$. The solution 
then transitions permanently to inference mode, with 
no further learning taking place. Formally, 
$\solution{}$ goes through three configurations:
\begin{itemize}
    \item For $\ti{} \in [\deploytime{}, \deploytime{} 
    + \adaptsetdim{} - 1]$, only $\feblock{}$ and 
    $\filterblock{}$ are executed, collecting incoming 
    instances into $\filterbatch{}^t$;
    \item at $\ti{} = \deploytime{} + \adaptsetdim{}$, 
    the trigger condition fires and $\learningmech{}$ 
    is executed on $\filterbatch{}^t$, producing a 
    trained $\algo{}^t$;
    \item for $\ti{} > \deploytime{} + \adaptsetdim{}$, 
    only $\feblock{}$ and $\algo{}^t$ are executed, 
    performing inference on $\stream{}$.
\end{itemize}
By construction, batch learning solutions have no 
mechanism for managing further changes in the 
distribution of $\dataprocess{}$ after the initial 
adaptation phase, and are therefore only applicable 
in the single-change regime. The surveyed batch 
learning solutions are summarized in 
Table~\ref{tab:batch_learning}, and the implementation of each component across these works is analyzed below.

\input{tables/batch_learning}

\subsubsection{Dimensionality Reduction Block 
$\feblock{}$}

The large majority of works implement $\feblock{}$ 
as a pre-trained Convolutional Neural Network (CNN) whose parameters remain frozen 
throughout the on-device 
phase~\cite{cereda_-device_2024, cereda_training_2024, 
cioflan_boosting_2024, cioflan_-device_2024, 
cioflan_towards_2022, craighero_-device_2024, 
pavan_tinysv_2025, rusci_-device_2023, 
rusci_few-shot_2023, rusci_self-learning_2025}. In 
audio applications, the CNN is typically preceded 
by a Mel-frequency cepstral coefficients (MFCC) front-end that converts the raw audio 
signal into a compact spectral representation before 
feature extraction~\cite{cioflan_boosting_2024, 
cioflan_-device_2024, cioflan_towards_2022, 
pavan_tinysv_2025, rusci_-device_2023, 
rusci_few-shot_2023, rusci_self-learning_2025}. A 
smaller group of works foregoes a learned feature 
extractor entirely: \cite{abdennadher_fixed_2021} 
relies on a fixed reservoir followed by Principal Component Analysis (PCA), while 
\cite{sudharsan_edge2train_2020, sudharsan_ml-mcu_2022, 
pereira_-device_2023} pass raw or minimally processed 
data directly to $\algo{}^t$. The works focusing on 
backpropagation 
optimization~\cite{cai_tinytl_2021, khan_dacapo_2023, 
lin_-device_2024, rub_tinyprop_2023, rub_advancing_2024, 
kwon_tinytrain_2024, rub_sparse_2025, 
deutel_-device_2025} set $\feblock{}$ to the identity 
function, incorporating the full network into 
$\algo{}^t$ instead. This last pattern is exclusive 
to method-oriented works: no application-oriented 
work in the batch learning literature trains end-to-end 
on-device, reflecting the current practical infeasibility of 
full network training on constrained hardware for 
real deployment scenarios.

\subsubsection{Data Management Block $\filterblock{}$}

Across all batch learning solutions, $\filterblock{}$ 
reduces to a buffer of fixed capacity $|\filterbatch{}|$ 
with a trivial admission condition: every incoming 
instance is stored until the buffer is full, at which 
point the trigger condition fires and $\learningmech{}$ 
is activated. The primary source of variation is in 
how $|\filterbatch{}|$ is determined. The large 
majority of works assume the complete adaptation 
dataset fits in memory and set $|\filterbatch{}|$ 
accordingly~\cite{abdennadher_fixed_2021, 
cai_tinytl_2021, cereda_-device_2024, 
cereda_training_2024, cioflan_boosting_2024, 
cioflan_-device_2024, cioflan_towards_2022, 
craighero_-device_2024, deutel_-device_2025, 
khan_dacapo_2023, lin_-device_2024, 
pereira_-device_2023, rub_advancing_2024, 
rub_sparse_2025, rub_tinyprop_2023, 
sudharsan_edge2train_2020, sudharsan_ml-mcu_2022}. 
A smaller group targets few-shot settings, fixing 
$|\filterbatch{}|$ to a small number of labeled 
examples per class, typically between one and 
five~\cite{pavan_tinysv_2025, rusci_-device_2023, 
rusci_few-shot_2023}. \cite{rusci_self-learning_2025} 
occupies an intermediate position, maintaining a 
balanced buffer of positive and negative examples 
whose size is determined by the self-labeling 
procedure rather than fixed in advance. Similarly, 
\cite{cioflan_boosting_2024, cioflan_-device_2024, 
cioflan_towards_2022} retain a portion of 
$\trainset{}$ in the buffer alongside new examples, 
anticipating the replay strategies seen in the continual learning regime.

The memory footprint of $\filterblock{}$, i.e., 
$\memory{}_{\filterblock{}}^{\ti{}} = |\filterbatch{}| 
\cdot |\latentrep{}^{\ti{}}|$, is frequently the 
largest term in Equation~\ref{eq:memory} and is 
the primary source of underestimation in works that 
do not target a real deployment device or implement 
a simplified version of the solution on-device. It 
is usually not reported in methodological works 
focused solely on backpropagation 
optimization~\cite{cai_tinytl_2021, lin_-device_2024, 
rub_advancing_2024, rub_sparse_2025, rub_tinyprop_2023, 
kwon_tinytrain_2024}, whose reported memory figures usually compare just the cost of $\algo^t$ and $\learningmech$ with and without the optimization.

% ── L + f ──
\subsubsection{ML Algorithm $\algo{}^t$ and Learning 
Mechanism $\learningmech{}$}

The joint choice of $\algo{}^t$ and $\learningmech{}$ 
can be organized into three families, which broadly 
correspond to the choice of $\feblock{}$ described 
above.

The first and most common family pairs a small MLP 
as $\algo{}^t$ with standard backpropagation as 
$\learningmech{}$. This combination is natural when 
$\feblock{}$ is a frozen pre-trained CNN: the 
learning phase is restricted to the classification 
head, keeping both the memory requirement of 
$\learningmech{}$ and the execution time 
$\executiontime{}_{\learningmech{}}^{\ti{}}$ 
minimal~\cite{cereda_-device_2024, cereda_training_2024, 
cioflan_boosting_2024, cioflan_-device_2024, 
cioflan_towards_2022, craighero_-device_2024}.

The second family uses a complete neural network as 
$\algo{}^t$, with $\feblock{}$ reducing to the 
identity function. $\algo{}^t$ is initialized from 
a network pre-trained on $\trainset{}$ and then 
partially updated on-device, with the associated 
increase in memory and execution time addressed 
through an optimized $\learningmech{}$. Approaches 
include bias-only 
updates~\cite{cai_tinytl_2021}, sparse parameter 
updates applied selectively to the most important 
parameters or 
channels~\cite{khan_dacapo_2023, lin_-device_2024, 
rub_tinyprop_2023, rub_advancing_2024, 
kwon_tinytrain_2024}, and forward-only learning 
rules that avoid backpropagation 
entirely~\cite{deutel_-device_2025, rub_sparse_2025}.

The third family foregoes neural networks entirely, 
pairing non-deep classifiers as $\algo{}^t$ with 
learning mechanisms tailored to their structure. 
Prototype-based classifiers are learned by 
constructing class representatives from averaged 
feature 
vectors~\cite{rusci_-device_2023, rusci_few-shot_2023, 
rusci_self-learning_2025}, instance-based classifiers 
store feature vectors directly for similarity-based 
classification~\cite{pavan_tinysv_2025}, SVMs are 
trained through their standard optimization 
procedure~\cite{abdennadher_fixed_2021, 
sudharsan_edge2train_2020}, linear classifiers are 
fitted through SGD with a one-vs-one 
scheme~\cite{sudharsan_ml-mcu_2022}, and parametric 
models are fitted through expectation-maximization~\cite{pereira_-device_2023}. 
As noted above, many of these solutions still rely 
on a frozen pre-trained CNN as $\feblock{}$, so 
that the non-deep nature applies to $\algo{}^t$ 
and $\learningmech{}$ only.

\subsection{Incremental Learning Solutions}
\label{sect:technical-il}

\begin{figure}[t]
    \centering
    \includegraphics[width=0.5\linewidth]{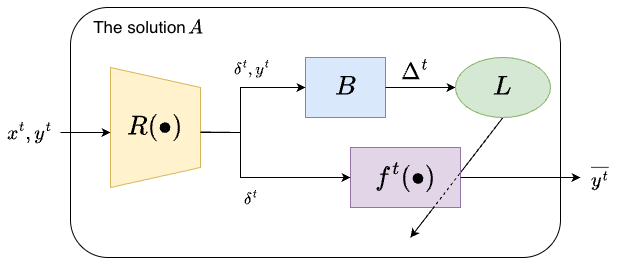}
    \Description{Diagram showing the temporal configuration of an incremental learning ODL solution, where all four components remain active throughout the device lifetime as the learning mechanism is executed repeatedly over the incoming data stream.}
    \caption{The configuration of an incremental
    learning solution $\solution{}$. Colors indicate that the component is active at time $t$.}
    \label{fig:continuous}
\end{figure}

In incremental learning solutions, $\learningmech{}$ 
is executed multiple times as $\stream{}$ progresses, 
updating $\algo{}^t$ every time the trigger condition 
is satisfied, as illustrated in 
Figure~\ref{fig:continuous}. All four components of $\solution{}$ 
remain active throughout the operational life of 
$\device{}$, though not all are required to execute 
at every time instant $\ti{}$. Unlike batch learning 
solutions, incremental learning solutions are 
applicable in all three regimes, as they retain the 
ability to adapt to further changes in the 
distribution of $\dataprocess{}$ throughout the 
operational life of $\device{}$. The surveyed 
incremental learning solutions are summarized in 
Table~\ref{tab:incremental_learning}, and their technical 
composition is analyzed below, organized by 
component.

%% ─── TABLE: INCREMENTAL LEARNING ─────────────────────────────────────────────
\input{tables/incremental_learning}

\subsubsection{Dimensionality Reduction Block 
$\feblock{}$}

In Incremental Learning, a frozen pre-trained CNN is also the dominant choice for $\feblock{}$ across all three 
regimes~\cite{ren_tinyol_2021, ren_-device_2024, 
disabato_incremental_2020, pau_tinyrce_2023, 
pau_tinyrce_2023-1, pavan_tactile_2025, 
rub_drip_2025, burrello_tackling_2021, 
kwon_lifelearner_2023, mei_train--request_2024, 
mei_ultra-low_2025, ravaglia_tinyml_2021, 
vorabbi_enabling_2024, wibowo_12_2024, 
disabato_tiny_2024, pavan_tybox_2024}. This pattern 
is driven by the constraints of $\device{}$ rather 
than algorithmic preference: by keeping $\feblock{}$ frozen, both 
$\memory{}_{\learningmech{}}^{\ti{}}$ and 
$\executiontime{}_{\learningmech{}}^{\ti{}}$ are kept 
minimal regardless of the regime. In the 
continual learning regime, this choice is nearly 
universal, with only \cite{rub_continual_2024} and 
\cite{leite_resource-efficient_2022} setting 
$\feblock{}$ to the identity and training end-to-end 
on-device. In \cite{kwon_lifelearner_2023} and 
\cite{wibowo_12_2024}, $\feblock{}$ is additionally 
meta-learned offline to produce representations 
particularly amenable to few-shot on-device 
adaptation.

The concept drift regime is the exception to this 
pattern. The majority of concept drift works set 
$\feblock{}$ to the identity function, operating 
directly on raw or minimally processed sensor 
data~\cite{pau_learning_2025, saccani_-sensor_2024, 
lourenco_dfdt_2025, ma_-demand_2025}, with only 
\cite{disabato_tiny_2024} and \cite{pavan_tybox_2024} 
relying on a frozen pre-trained CNN. This reflects 
the nature of the tasks addressed in that regime: 
sensor calibration and data stream classification 
operate on low-dimensional tabular or scalar signals 
that do not require a learned feature extractor, 
unlike the audio and image inputs that dominate the 
single-change and continual learning regimes.

\subsubsection{Data Management Block $\filterblock{}$}

The configuration of $\filterblock{}$ reveals the 
sharpest cross-regime divergence in the incremental 
learning literature. In the single-change and concept 
drift regimes, fully online operation with 
$|\filterbatch{}| = 1$ is the dominant strategy, 
meaning $\learningmech{}$ is executed on each 
incoming instance 
individually~\cite{ren_tinyol_2021, ren_-device_2024, 
disabato_incremental_2020, chowdhary_-sensor_2023, 
zanghieri_semg-driven_2024, renzi_online_2025, 
sudharsan_train_2021, pau_learning_2025, 
saccani_-sensor_2024, lourenco_dfdt_2025}. This 
choice minimizes the memory footprint of 
$\filterblock{}$, keeping 
$\memory{}_{\filterblock{}}^{\ti{}} \approx 0$, and 
is consistent with the lightweight algorithmic 
families that dominate these regimes. A smaller 
group of single-change works uses explicit buffering: 
\cite{pavan_tactile_2025} and \cite{rub_drip_2025} 
adopt an active learning strategy that selects only the most informative instances to add to the buffer; 
\cite{pau_tinyrce_2023, pau_tinyrce_2023-1} empty 
the buffer after each update; while 
\cite{burrello_tackling_2021} maintains a 
fixed-capacity buffer with uniform and exponential 
sampling strategies. In the concept drift regime, 
\cite{pavan_tybox_2024} maintains a First-in-First-out (FIFO) buffer of 
fixed capacity, and \cite{disabato_tiny_2024} and 
\cite{ma_-demand_2025} activate $\learningmech{}$ 
only when an explicit change detection mechanism 
signals a distribution shift.

The continual learning regime stands in sharp 
contrast: all solutions use explicit buffering, 
with $\filterblock{}$ collecting instances 
corresponding to a new distribution before 
triggering $\learningmech{}$. This is a structural 
necessity rather than a design preference, as 
preventing catastrophic forgetting requires 
retaining some representation of past distributions 
alongside new data, in the most common continual learning solutions. The primary sources of variation 
are in how the buffer content is managed and when 
$\learningmech{}$ is triggered. To minimize the 
memory footprint of the replay buffer, 
\cite{ravaglia_tinyml_2021} and 
\cite{vorabbi_enabling_2024} store compact latent 
representations of past distributions using 
quantized and binary encodings respectively, while 
\cite{rub_continual_2024} populates 
$\filterbatch{}^t$ with synthetic samples obtained 
through dataset distillation rather than real 
instances. \cite{kwon_lifelearner_2023} reduces the 
buffer requirements from the other direction, using 
a meta-learned $\feblock{}$ that produces 
representations amenable to few-shot adaptation, 
so that fewer past samples are needed to anchor 
previous knowledge. Regarding the trigger condition, 
\cite{mei_train--request_2024} activates 
$\learningmech{}$ only when classification accuracy 
drops below an acceptability threshold, while all other works trigger when the buffer is full.

\subsubsection{ML Algorithm $\algo{}^t$ and Learning 
Mechanism $\learningmech{}$}

The joint choice of $\algo{}^t$ and $\learningmech{}$ 
follows a clear cross-regime pattern. In the 
continual learning regime, backpropagation applied 
to a DNN is the dominant and nearly universal 
combination~\cite{kwon_lifelearner_2023, 
mei_train--request_2024, mei_ultra-low_2025, 
ravaglia_tinyml_2021, vorabbi_enabling_2024, 
rub_continual_2024, leite_resource-efficient_2022}, 
with \cite{wibowo_12_2024} as the only exception, 
using a prototype-based classifier whose 
$\learningmech{}$ reduces to constructing a new 
class representative from averaged feature vectors. 
The structural reason is straightforward: since all continual learning solutions maintain a replay buffer of sufficient size to prevent catastrophic forgetting, the memory cost of backpropagation is unlikely to constitute a bottleneck relative to the buffer itself.

In the single-change and concept drift regimes, the 
picture is more diverse. Backpropagation-based DNNs 
remain present in both: \cite{ren_tinyol_2021, 
ren_-device_2024, rub_drip_2025, pavan_tactile_2025, 
burrello_tackling_2021, zanghieri_semg-driven_2024} 
in single-change, and \cite{pavan_tybox_2024} and 
\cite{ma_-demand_2025} in concept drift. However, 
lighter alternatives are well represented. 
Forward-only rules appear in both regimes: 
\cite{pau_tinyrce_2023, pau_tinyrce_2023-1} pair a 
hyperspherical Restricted Coulomb Energy (RCE) classifier with a forward-only 
update in single-change, while \cite{pau_learning_2025} 
and \cite{saccani_-sensor_2024} use Radial Basis Function (RBF) networks 
with a forward-only mechanism in concept drift. 
Non-neural approaches are also present: K Nearest Neighbors (KNN)-based 
classifiers in \cite{disabato_incremental_2020} and 
\cite{disabato_tiny_2024}, prototype construction 
in \cite{chowdhary_-sensor_2023}, a binary linear 
classifier in \cite{sudharsan_train_2021}, a 
Hoeffding-inspired decision tree in 
\cite{lourenco_dfdt_2025}, and a multi-classifier 
evaluation in \cite{renzi_online_2025}.

The concentration of backpropagation in continual 
learning and its diversification in the other two 
regimes reflects a structural property discussed 
further in Section~\ref{sect:technical-summary}: 
instance-based and prototype-based solutions tend to grow monotonically in memory over time (at least in their basic version without optimizations), 
making them poorly suited to regimes where 
$\solution{}$ must remain active and adapt 
indefinitely.

\subsection{Findings}
\label{sect:technical-summary}

\subsubsection{Solution Components, Application Environments, and Hardware classes}

\input{tables/components}

The patterns observed across the surveyed solutions 
can be understood as consequences of the trade-offs 
inherent in each component choice, with hardware 
capability acting as the primary constraint shaping 
which combinations are feasible. Table~\ref{tab:components} summarizes these 
trade-offs with qualitative estimates of $\memory{}_{\solution{}}^{\ti{}}$ and $\executiontime{}(\solution{}^{\ti{}}, \device{})$ for each component choice. 
The estimates are necessarily qualitative rather 
than quantitative, for a reason that goes beyond 
measurement difficulty: no component can be 
evaluated in isolation from the others. A frozen 
pre-trained CNN as $\feblock{}$ produces compact 
$\latentrep{}^t$ that allow $\algo{}$ to be a 
simple classifier, whereas identity forces $\algo{}$ 
to absorb the full input complexity. Similarly, 
full backpropagation may be viable on an 
ultra-constrained device if $\algo{}$ is a single 
linear layer, yet infeasible if it is a deep CNN. 
The table should therefore be read as a 
characterization of the trade-offs each choice 
introduces, not as an independent ranking of 
alternatives.

For $\feblock{}$, the choice ranges from identity 
(no overhead, but full input dimensionality exposed 
to $\learningmech{}$) through statistical feature 
extractors (lightweight dimensionality reduction 
without pre-training) to frozen pre-trained CNNs 
(highest cost, but compact discriminative 
representations that directly reduce 
$|\latentrep{}^t|$, the memory footprint of 
$\filterbatch{}^t$, and the labeled data required 
for adaptation). The frozen CNN is essentially 
mandatory for image and audio inputs, and is the 
dominant choice on standard MCUs and PULP platforms. 
Identity and statistical extractors are more common 
on ultra-constrained devices, where the cost of 
running a CNN may be prohibitive, and where simpler 
signal types are typically used.

For $\filterblock{}$, fully online operation with 
$|\filterbatch{}| = 1$ minimizes 
$\memory{}_{\filterblock{}}^{\ti{}}$ and per-update 
execution time, but limits $\learningmech{}$ to a 
single pass per instance, making a larger number of 
labeled instances necessary to reach convergence. Explicit buffering with 
$|\filterbatch{}| > 1$ recovers data efficiency at 
the cost of a larger $\memory{}_{\filterblock{}}^{\ti{}}$. 
Batch collection is the extreme case: peak RAM is largest, as $\memory{}_{\filterblock{}}^{\ti{}}$ and the requirements of $\learningmech{}$ and $\algo{}^t$ overlap during the single learning episode, after which the solution runs inference only. This cost can nonetheless be kept within budget when the learning episode is designed for few-shot settings, where $|\filterbatch{}|$ remains small by construction. Online 
solutions dominate in single-change and concept 
drift regimes. Explicit buffering is structurally 
necessary in continual learning, where past data 
must be retained to prevent catastrophic forgetting.

For $\algo{}^t$ and $\learningmech{}$, non-deep 
classifiers are cheaper to update than 
backpropagation-based networks, but instance-based 
and prototype-based methods grow in memory as new 
data are incorporated. Neural networks have a fixed 
memory footprint regardless of data seen, making 
them better suited to non-stationary regimes. Within 
neural network solutions, structured architectures 
such as RBF and RCE networks reduce learning to a 
forward pass, while sparsification, quantization, 
and binarization lower the cost of backpropagation. 
These optimizations concentrate in the single-change 
regime, where per-update cost is the primary 
concern. In continual learning, full backpropagation 
on capable hardware is the dominant choice, driven 
by the explicit replay buffers that all solutions 
in that regime require.

\subsubsection{The Gap Between Methodological and 
Application-Oriented Solutions}

Two structural gaps emerge when comparing methodological 
and application-oriented works. 

The first concerns 
scope. Backpropagation optimization, the largest single 
cluster of methodological contributions in this survey, 
has produced a rich body of work on sparsification, 
quantization, and forward-only alternatives, yet most 
of these contributions are either validated only on 
standard benchmarks without any on-device implementation, 
or deployed on standard MCUs under controlled conditions 
that do not reflect real application constraints. 
Either way, this work has not translated into application 
deployments: no application-oriented work across any 
of the three regimes adopts these optimizations. The 
reason is not technical incompatibility but a narrowness 
of focus: methodological works optimize a single 
component of the pipeline in isolation, typically 
$\learningmech{}$, and measure only its cost, without 
considering that the bottleneck in a real deployment 
may lie elsewhere, for instance in $\filterblock{}$ 
or in the labeling strategy. The ODL 
problem is not solved by optimizing one component: 
it requires the entire pipeline to be viable 
simultaneously.

The second gap concerns label availability. 
Methodological works uniformly assume that labeled 
data is available in the quantities needed to validate 
their approach, an assumption that is rarely satisfied 
in practice, and which is particularly acute in the 
concept drift and continual learning regimes where 
application-oriented works remain almost entirely 
absent. Label availability is often the binding 
constraint that prevents methodological solutions from 
being ported to real deployments, and backpropagation 
optimization and label-efficient learning are rarely 
developed together despite being complementary concerns.
The application-oriented works that have succeeded 
in real deployments have done so precisely because 
they treated label acquisition as a first-class design 
requirement alongside the algorithmic and hardware 
choices. Four distinct strategies emerge from these 
works. \textit{Few-shot learning} provides adaptation 
from a small number of labeled examples supplied by 
the user, and is adopted 
by~\cite{rusci_-device_2023, rusci_self-learning_2025, 
pavan_tinysv_2025, cioflan_-device_2024} for 
personalization tasks such as keyword enrollment and 
speaker verification. \textit{Self-supervised learning 
with an auxiliary sensor} derives labels automatically 
from a secondary on-board sensor, eliminating human 
annotation entirely, as 
in~\cite{cereda_-device_2024, cereda_training_2024} 
where supervision comes from the drone's flight 
controller during adaptation. \textit{Unsupervised 
learning} requires no labels at all, and applies 
naturally to anomaly detection and sensor calibration, 
where normality is learned from unlabeled 
post-deployment 
data~\cite{pereira_-device_2023, ren_tinyol_2021}. 
\textit{Implicit labeling through structured stimulus 
sequences} derives labels from a predefined enrollment 
protocol rather than explicit annotation, as adopted 
by~\cite{chowdhary_-sensor_2023, mei_ultra-low_2025} 
for biosignal applications. Each strategy directly 
constrains the viable solution architecture: few-shot 
works pair a frozen CNN with a prototype or 
instance-based classifier, unsupervised works avoid 
backpropagation entirely, and implicit labeling is 
specific to settings where an enrollment phase can 
be designed into the deployment protocol.

Taken together, both gaps point toward the same 
underlying need: solutions and benchmarks that treat 
the full pipeline as the unit of design and evaluation, 
rather than optimizing and evaluating individual components in 
isolation. This means end-to-end frameworks that expose 
$\feblock{}$, $\filterblock{}$, $\algo{}$, and 
$\learningmech{}$ as jointly optimizable objects, and 
application-oriented benchmarks that reflect realistic 
labeling conditions and deployment constraints, 
as discussed further in Section~\ref{sect:future}.

\subsubsection{Solution Reporting}

\input{tables/reporting}

Evaluating and comparing ODL solutions is 
substantially harder than evaluating static ML 
models. The difficulty manifests across two classes 
of metrics: joint metrics that depend on both 
$\solution{}$ and $\device{}$, and solution-only 
metrics that characterize $\solution{}$ independently 
of hardware.

\paragraph{Joint metrics: latency and energy.}
Table~\ref{tab:reporting} summarizes the 
normalization conventions used for latency and 
energy across the surveyed works. The fragmentation 
is striking: figures are reported per instance, per 
batch, per epoch, per learning event, and per 
complete training run, with no consistency across 
works or even within the same regime. This makes 
direct comparison impossible even between solutions 
targeting the same hardware and task. The problem 
is not merely cosmetic: a latency reported per 
complete training run and one reported per instance 
differ by orders of magnitude for the same solution, 
depending on dataset size.

Regime-level patterns are visible in
Table~\ref{tab:reporting}. In the single-change
regime, reporting per complete training run dominates
for both latency and energy: the learning phase is a
one-off event, so total cost is the most operationally
relevant figure. In the concept drift regime, energy
is entirely absent from the reported metrics, a
critical gap given that the cumulative cost of
continuously repeating small updates is precisely what
matters in this regime; per-instance or per-update
figures would be the natural unit yet are rarely used.
In the continual learning regime, ``per learning
event" is the most principled unit, reflecting the
episodic structure of task-incremental updates, and
the works that do report energy adopt it
consistently~\cite{ravaglia_tinyml_2021,
kwon_lifelearner_2023}.

In summary, joint metrics are present across the literature but reported with inconsistent normalization conventions, making direct cross-work comparison unreliable.

\paragraph{Solution-only metrics: memory, 
computational cost, and predictive performance.}

Peak RAM is the most consistently reported metric, 
yet many works report only the memory footprint of 
$\learningmech{}$ and $\algo{}^t$, ignoring 
$\memory{}^t_{\filterblock{}}$, which as argued 
above can be the dominant term in 
Equation~\ref{eq:memory}. While most surveyed works report hardware-dependent metrics such as $\executiontime{}\!\left(\solution{}^{\ti{}}, \device{}\right)$ 
or the memory budget $\memorydevice{}$, a minority resort to hardware-independent proxies such as 
FLOPs, multiply-accumulate operations (MACs), parameter counts, or activation 
sizes~\cite{cai_tinytl_2021, kwon_tinytrain_2024, rub_sparse_2025, saccani_-sensor_2024, rub_continual_2024, vorabbi_enabling_2024, renzi_online_2025}, 
typically when no deployment target $\device{}$ is available. Although these metrics 
do not directly capture the peak RAM memory consumption or the execution time on a real $\device{}$, 
their hardware independence is a genuine advantage: a worst-case analytical estimate 
can, for certain classes of solutions, enable fairer cross-platform algorithmic 
comparison than wall-clock measurements tied to a specific $\device{}$.

On the predictive performance side, the time axis 
introduced by ODL makes comparison fundamentally 
harder than in static ML. In non-stationary regimes, 
a single accuracy figure is insufficient: a solution 
that adapts quickly but forgets rapidly is not 
comparable to one that adapts slowly but retains 
past knowledge. Even in the single-change regime, 
the amount of adaptation data is a critical variable 
that is rarely controlled for: works report results 
at vastly different numbers of labeled examples per 
class, making cross-paper comparison unreliable. 
This is discussed further in 
Section~\ref{sect:future}.

\section{Future Directions}
\label{sect:future}

The analyses carried out in this survey point toward 
a coherent set of open challenges that, taken 
together, suggest a broader research agenda for 
the field.

\paragraph{Standardized evaluation frameworks.}
A fundamental obstacle to progress in ODL is the 
absence of standardized evaluation frameworks. As 
highlighted in Sections~\ref{sect:application-summary}, 
\ref{sect:hardware-summary}, and 
\ref{sect:technical-summary}, the literature 
currently lacks consistent evaluation conventions 
at every level of the analysis. Three complementary 
frameworks are needed, each addressing a different 
aspect of the problem.

A \textit{solution-oriented benchmark}, aimed at evaluating $\solution{}$ independently of hardware, should fix a set of worst-case RAM and computational budgets and report predictive performance for solutions designed within each budget. Results should span multiple stream lengths to capture the adaptation curve rather than final accuracy alone, and the full memory footprint $\memory{}_{\solution{}}^{\ti{}}$, including $\memory{}_{\filterblock{}}^{\ti{}}$, must be verified against each budget. Where possible, deploying all candidate solutions on a common reference platform would allow joint hardware-solution metrics to be collected and worst-case estimates to be confirmed empirically. This benchmark is most directly applicable to method-oriented works, where the goal is to demonstrate algorithmic effectiveness across a range of resource constraints.

A \textit{hardware-oriented benchmark}, aimed at evaluating the ODL capabilities of a given $\device{}$, should fix a set of ODL solutions and report the joint hardware-solution metrics: latency $\executiontime{}\!\left(\solution{}^{\ti{}}, \device{}\right)$, energy consumption for a clearly defined learning event, and peak instantaneous active power. The relevant hardware specifications, including $\memorydevice{}$ and $\frequency{}$, should be reported alongside to enable fair cross-platform comparison. This enables direct, reproducible comparison of hardware platforms independently of algorithmic choices, as argued in Section~\ref{sect:hardware-summary}. This benchmark is most directly applicable to the evaluation of novel hardware platforms, whether research prototypes or commercial devices, across a range of ODL tasks and solutions.

A \textit{co-design benchmark}, aimed at characterizing the tradeoff between $\solution{}$ and $\device{}$, should specify explicit budgets for latency $\executionconstraint{}$ and power, and report joint hardware-solution metrics (latency $\executiontime{}\!\left(\solution{}^{\ti{}}, \device{}\right)$ and energy per learning event) together with predictive performance over time, after an optimization process that may involve changes to both $\solution{}$ and $\device{}$. As argued in Section~\ref{sect:application-summary}, the absence of such benchmarks currently prevents co-optimization in application-oriented works, forcing hardware and algorithmic choices to be made sequentially rather than jointly. This benchmark is most directly applicable to application-oriented works, where constraints are dictated by application requirements and the priority is to find the best tradeoff between solution complexity and hardware cost.

All three benchmarks should be organized by 
distribution change regime, since solutions 
operating under different regimes are not directly 
comparable, and should be accompanied by clear 
instructions for dataset construction to enable 
the community to extend them to new application 
domains. Real-world application-oriented works, 
such as those surveyed in 
Section~\ref{sect:application-real}, provide the 
most concrete inspiration for what these benchmarks 
should look like in practice.

Addressing label efficiency should be an explicit 
design requirement of the solution-oriented 
benchmark rather than a separate concern. Concretely, 
this means constructing data streams that mix 
labeled and unlabeled instances in realistic 
proportions, and evaluating solutions on their 
ability to exploit unlabeled data alongside the 
scarce labeled examples they receive. Such 
benchmarks, particularly if designed for the 
concept drift and continual learning regimes where 
application-oriented works are currently almost 
absent, would reflect the actual conditions under 
which ODL solutions must operate in the real world. 
This in turn would create the conditions for 
developing solutions in these regimes that can be 
transferred to real deployments with greater 
confidence, rather than remaining validated only 
on synthetic or laboratory data streams.

\paragraph{End-to-end design of solution components.}
As argued in Section~\ref{sect:technical-summary}, 
the surveyed literature predominantly optimizes 
components of $\solution{}$ in isolation, yet the 
overall efficiency of $\solution{}$ depends on all 
four components jointly. Closing this gap requires 
not just better benchmarks, but dedicated software 
tools: libraries and design frameworks that expose 
all components of $\solution{}$ as jointly 
optimizable objects, support end-to-end evaluation 
under realistic deployment constraints, and make 
the cost of each component visible during the 
design process. No such tool currently exists in 
the ODL literature, and its development would 
create the infrastructure needed for principled 
end-to-end design of $\feblock{}$, $\filterblock{}$, 
$\algo{}$, and $\learningmech{}$ jointly.

\paragraph{Hardware for ODL}
As argued in Section~\ref{sect:hardware-summary}, 
the dominant architectural pattern in the surveyed 
literature keeps the heaviest computation in the 
frozen $\feblock{}$ inference pass. Hardware that 
accelerates inference efficiently is therefore 
currently more impactful for ODL than hardware 
that accelerates gradient computation.

Looking further ahead, the constraints of ODL 
differ fundamentally from those of cloud or edge 
training: labeled data is scarce, validation 
cannot happen on-device, power budgets are tight, 
and the solution must remain operational throughout 
the learning process. A hardware platform designed 
with these constraints in mind from the outset, 
rather than adapted from cloud computing systems architecture, 
could offer qualitatively different trade-offs. 
Whether such platforms emerge from extensions of 
current MCU architectures or from entirely new 
design paradigms remains an open question.

\section{Conclusions}
\label{sect:conclusions}

This survey has presented a comprehensive analysis of 
On-Device Learning for TinyML, organized around a 
single unifying principle: the distribution change 
regime that a solution is designed to operate under. 
By distinguishing between single-change, concept drift, 
and continual learning regimes, and analyzing each 
through application, hardware, and solution lenses, 
the survey provides a framework for comparing ODL 
solutions in a principled way that prior surveys have 
not adopted.

From the application lens, a significant structural 
gap emerges between methodological and 
application-oriented works. Methodological works 
cluster around image classification benchmarks too 
demanding for most real-world TinyML deployments, 
while the domains that do fit within TinyML 
constraints lack standardized benchmarks for fair 
comparison. The works that have succeeded in real 
deployments share a common characteristic: they treat 
label acquisition as a first-class design requirement, 
through few-shot learning, self-supervision, 
unsupervised approaches, or implicit labeling. 
Application-oriented works in the concept drift and 
continual learning regimes remain almost entirely 
absent, reflecting both the difficulty of obtaining 
labeled data in non-stationary settings and the 
absence of realistic benchmarks for these regimes.

From the hardware lens, a clear monotonic correlation 
emerges between hardware capability and regime 
complexity: ultra-constrained processors appear almost 
exclusively in the single-change regime, standard MCUs 
span all three but are thinly represented in continual 
learning, and PULP platforms dominate where memory 
demands are largest. Across the literature, the 
heaviest computation consistently resides in the frozen 
feature extractor inference pass, meaning hardware that 
accelerates inference efficiently is currently more 
impactful for ODL than hardware designed to accelerate 
gradient computation.

From the solution lens, the combination of a frozen 
pre-trained feature extractor and a lightweight 
learnable classifier is the dominant pattern across technical compositions in
all three regimes. Backpropagation optimization 
concentrates in the single-change regime, while 
non-stationary regimes rely more heavily on 
prototype-based classifiers, forward-only learning, 
and latent replay. A recurring limitation is that 
components are optimized in isolation, yet the overall 
efficiency of a solution depends on all four components 
jointly, and them in an end-to-end manner under realistic 
deployment constraints remains underexplored.

Taken together, these findings point toward a coherent research agenda with three open challenges. First, the field lacks standardized evaluation frameworks: a solution-oriented benchmark evaluating $\solution{}$ across RAM and computational budgets, a hardware-oriented benchmark comparing $\device{}$ on a fixed solution, and a co-design benchmark jointly optimizing $\solution{}$ and $\device{}$ under explicit deployment constraints. Second, dedicated software tools are needed that expose $\feblock{}$, $\filterblock{}$, $\algo{}$, and $\learningmech{}$ as jointly optimizable objects, making end-to-end co-design of ODL solutions tractable. Third, hardware designed specifically for ODL constraints (scarce labels, no on-device validation, tight power budgets), rather than adapted from cloud training architectures, could offer qualitatively different tradeoffs than current MCU-class platforms. The framework introduced in this survey, by making the distributional assumptions of each solution explicit and enabling meaningful comparison across the growing body of ODL literature, provides a foundation for addressing these challenges.

\bibliographystyle{ACM-Reference-Format}
\bibliography{biblio}

\begin{acks}
The authors used Claude Sonnet 4.6 (Anthropic) as an AI-assisted writing tool to support text editing and revision during the preparation of this manuscript. All outputs were reviewed, revised, and validated by the authors, who take full responsibility for the content.
\end{acks}

\section{Mathematical Notation}
\label{app:notation}

Table~\ref{tab:notation} lists all mathematical symbols used throughout the paper.

\input{tables/notation}

\end{document}

%% file: Nomenclature.tex
%=================== Problem formulation ===============

\newcommand{\solution}{A}
\newcommand{\dataprocess}{P}
\newcommand{\device}{D}
\newcommand{\ti}{t}
\newcommand{\classes}{\Gamma}
\newcommand{\deploytime}{T}
\newcommand{\trainset}{S_0}
\newcommand{\testset}{S_{test}}
\newcommand{\stream}{S}
\newcommand{\memory}{M}
\newcommand{\executiontime}{E}
\newcommand{\op}{Op}
\newcommand{\metric}{m}
\newcommand{\frequency}{F_{\device{}}}
\newcommand{\executionconstraint}{\overline{\executiontime{}}}
\newcommand{\memorydevice}{\memory{}_{\device{}}}

\newcommand{\inp}{x^t}

%===================== Proposed Solution ==================
\newcommand{\fe}{R}
\newcommand{\feblock}{\fe}
\newcommand{\latentrep}{\delta}
\newcommand{\filterbatch}{\Delta}
\newcommand{\filterblock}{B}
\newcommand{\algo}{f}
\newcommand{\learningmech}{L}
\newcommand{\pred}{\overline{y^t}}
\newcommand{\adaptsetdim}{N}

%% file: tables/survey_comparison.tex
% Table: survey comparison (tab:survey_comparison) — paste content here
\begin{table*}[t]
\centering
\caption{Comparison of related surveys on on-device learning for TinyML.
  Taxonomy abbreviations: \textbf{NN} = Neural Networks, \textbf{DT} = Decision Trees.
  Aspects column: \textbf{M} = Models, \textbf{A} = Application, \textbf{HW} = Hardware.}
\label{tab:survey_comparison}
\resizebox{\textwidth}{!}{%
\begin{tabular}{llcccccc}
\toprule
\textbf{Survey} & \textbf{Year} & \textbf{Tiny Devices} & 
\textbf{Taxonomy} & \textbf{Aspects} & \textbf{\#ODL Works} \\
\midrule
Rajapakse et al.~\cite{rajapakse_intelligence_2023}  
  & 2023 & \cmark & offline v online & M, A          & $\sim$10 \\
Abozaid et al.~\cite{abozaid_adaptive_2025}      
  & 2025 & \cmark &  --  & M         & $\sim$10 \\
Lourenço et al.~\cite{lourenco_-device_2025}    
  & 2025 & \xmark & NN v DT  & M         & $\sim$10 \\
\midrule
\textbf{This survey} & 2026 & \cmark & Distr. Change & M, A, HW & $\sim$70 \\
\bottomrule
\end{tabular}%
}
\end{table*}

%% file: tables/benchmarks.tex
% Table: benchmarks (tab:benchmarks) — paste content here

\begin{table*}[t]
\centering
\caption{Benchmark datasets used in method-oriented ODL 
works, organized by regime and task.}
\label{tab:benchmarks}
\resizebox{\textwidth}{!}{%
\begin{NiceTabular}{lllp{0.4\textwidth}}[
  code-before =
    \rectanglecolor{envcolor}{2-1}{23-1}
    \rectanglecolor{envcolor}{23-1}{27-1}
    \rectanglecolor{envcolor}{27-1}{34-1}
]
\toprule
\textbf{Regime} & \textbf{Task} & \textbf{Dataset} & 
\textbf{Used in} \\
\midrule
% ── Single Change ─────────────────────────────────────────────
\Block{11-1}{Single-Change}
  & \Block{11-1}{Image Classification}
  & CIFAR-10~\cite{krizhevsky2009CIFAR}  & \cite{cai_tinytl_2021, lin_-device_2024, 
    deutel_-device_2025, rub_tinyprop_2023, 
    rub_advancing_2024, rub_sparse_2025, rub_drip_2025} \\
& & CIFAR-100~\cite{krizhevsky2009CIFAR} & \cite{cai_tinytl_2021, lin_-device_2024, 
    rub_advancing_2024} \\
& & MNIST~\cite{mnist}  & \cite{deutel_-device_2025, 
    rub_tinyprop_2023, rub_advancing_2024, rub_sparse_2025, 
    disabato_incremental_2020, rub_drip_2025,
    sudharsan_edge2train_2020, sudharsan_ml-mcu_2022, 
    sudharsan_train_2021} \\
& & Flowers~\cite{Nilsback_2008_flowers}, Food~\cite{Bossard_2014_Food} & \cite{cai_tinytl_2021, lin_-device_2024, 
    rub_advancing_2024, rub_sparse_2025} \\
& & Cars~\cite{Krause2013cars}, CUB~\cite{Wah_2011_CUB_200}, Pets~\cite{Parkhi_2012_Pets} & \cite{cai_tinytl_2021, 
    lin_-device_2024} \\
& & Aircraft~\cite{maji2013aircraft}   & \cite{cai_tinytl_2021} \\
& & VWW~\cite{chowdhery_2019_Visual}       & \cite{lin_-device_2024} \\
& & MiniImageNet~\cite{Vinyals_2016_Mini-imagenet} & \cite{kwon_tinytrain_2024} \\
& & Plant Disease~\cite{Ruth_2022_PlantDisease} & \cite{rub_drip_2025} \\
& & Omniglot~\cite{Lake_2015_Omniglot} & \cite{ren_-device_2024} \\
& & ImageNet~\cite{Deng_2009_imagenet}  & \cite{disabato_incremental_2020} \\
\cmidrule{2-4}
& \Block{2-1}{Audio Classification}
  & GSC~\cite{warden_speech_2018}       & \cite{rub_advancing_2024, rub_sparse_2025, 
    pau_tinyrce_2023, rub_drip_2025, ren_-device_2024} \\
& & DCASE2020~\cite{heittola2020DCASE} & \cite{rub_tinyprop_2023, 
    rub_advancing_2024} \\
\cmidrule{2-4}
& \Block{2-1}{IMU-HAR}
  & UCI-HAR~\cite{Anguita2013UCIHAR}   & \cite{rub_advancing_2024, 
    rub_sparse_2025} \\
& & PAMAP2~\cite{Reiss2012PAMAP2}, SHL~\cite{Wang2024SHL}, CWRU~\cite{smith_2015_cwru_bearing_data} & \cite{pau_tinyrce_2023-1} \\
\cmidrule{2-4}
& \Block{2-1}{Tabular}
  & Iris~\cite{iris_53}      & \cite{sudharsan_edge2train_2020, 
    sudharsan_ml-mcu_2022, sudharsan_train_2021} \\
& & Heart Disease~\cite{heart_disease_45}, Cancer~\cite{breast_cancer_1993_wisconsin} & \cite{sudharsan_ml-mcu_2022, 
    sudharsan_train_2021} \\
\midrule
% ── Concept Drift ─────────────────────────────────────────────
\Block{4-1}{Concept Drift}
  & \Block{2-1}{Image Classification}
  & MNIST\cite{mnist, fashion-mnist}     & \cite{pavan_tybox_2024} \\
& & CIFAR-10-C~\cite{Hendrycks_2019_CIfar-C}, ImageNet-C~\cite{Hendrycks_2019_CIfar-C}, SHIFT~\cite{sun2022shiftsyntheticdrivingdataset} & \cite{ma_-demand_2025} \\
\cmidrule{2-4}
& Audio Classification
  & GSC~\cite{warden_speech_2018}       & \cite{disabato_tiny_2024} \\
\cmidrule{2-4}
& Data Stream
  & DataStream Benchmark Suite~\cite{Souza_2020_Stream} & 
    \cite{lourenco_dfdt_2025} \\
\midrule
% ── Continual Learning ────────────────────────────────────────
\Block{7-1}{Continual Learning}
  & \Block{5-1}{Image Classification}
  & CIFAR-10~\cite{krizhevsky2009CIFAR}& \cite{rub_continual_2024, 
    vorabbi_enabling_2024} \\
& & CIFAR-100~\cite{krizhevsky2009CIFAR} & \cite{kwon_lifelearner_2023, 
    wibowo_12_2024} \\
& & CORe50~\cite{lomonaco_2017_Core50}    & \cite{ravaglia_tinyml_2021, 
    rub_continual_2024, vorabbi_enabling_2024} \\
& & MiniImageNet~\cite{Vinyals_2016_Mini-imagenet} & \cite{kwon_lifelearner_2023} \\
& & MNIST~\cite{mnist}  & \cite{rub_continual_2024} \\
\cmidrule{2-4}
& Audio Classification
  & GSC~\cite{warden_speech_2018}       & \cite{kwon_lifelearner_2023, 
    rub_continual_2024} \\
\cmidrule{2-4}
& IMU-HAR
  & UCI-HAR~\cite{Anguita2013UCIHAR}   & \cite{rub_continual_2024} \\
\bottomrule
\end{NiceTabular}%
}
\end{table*}

%% file: tables/applications.tex
% Table: applications (tab:applications) — paste content here
\begin{table*}[t]
\centering
\caption{Application-oriented ODL works, organized by 
regime and task.}
\label{tab:applications}
\resizebox{\textwidth}{!}{%
\begin{NiceTabular}{llllp{0.2\textwidth}}[
  code-before =
    \rectanglecolor{envcolor}{2-1}{15-1}
    \rectanglecolor{envcolor}{16-1}{17-1}
    \rectanglecolor{envcolor}{18-1}{21-1}
]
\toprule
\textbf{Regime} & \textbf{Task} & \textbf{Specific Task} & 
\textbf{Works} & \textbf{Dataset} \\
\midrule
% ── Single Change ─────────────────────────────────────────────
\Block{15-1}{Single-Change}
  & \Block{4-1}{KWS and Audio Classification}
  & Speaker adaptation & \cite{cioflan_boosting_2024} & GSC~\cite{warden_speech_2018} \\
& & New keyword enrollment & \cite{rusci_-device_2023, 
    rusci_few-shot_2023, rusci_self-learning_2025, 
    rusci_self-incremental_2025, 
    chauhan_exploring_2022} & GSC~\cite{warden_speech_2018}, HeySnips~\cite{leroy2019federatedlearningkeywordspotting}, 
    HeySnapdragon~\cite{2019heysnapdragon}, MSWC~\cite{mazumder2021multilingual}, Collected \\
& & Noise adaptation & \cite{cioflan_towards_2022, 
    cioflan_-device_2024} & GSC~\cite{warden_speech_2018} \\
& & Speaker verification & \cite{pavan_tinysv_2025} & 
    GSC~\cite{warden_speech_2018}, Collected \\
\cmidrule{2-5}
& \Block{1-1}{Human Pose Estimation}
  & Image: environment adaptation & \cite{cereda_-device_2024, 
    cereda_training_2024} & Collected \\
\cmidrule{2-5}
& \Block{1-1}{Human Activity Recognition}
  & IMU: user adaptation & \cite{craighero_-device_2024} & 
    WISDM~\cite{2011Wisdm}, Collected \\
\cmidrule{2-5}
& \Block{3-1}{Biosignal Classification}
  & sEMG: gesture recognition & \cite{burrello_tackling_2021, 
    benatti_online_2019} & UniBo-20-Session~\cite{Zanghieri2020Unibo20}, Collected \\
& & sEMG: hand dynamics estimation & 
    \cite{zanghieri_semg-driven_2024} & HYS›ER~\cite{Hyser2021} \\
& & ECG: anomaly detection & 
    \cite{abdennadher_fixed_2021} & MIT-BIH~\cite{Moody_2001_MIT-BIH} \\
\cmidrule{2-5}
& \Block{3-1}{Anomaly Detection}
  & Industrial time series & \cite{pereira_-device_2023} & 
    Collected \\
& & IMU: Fan monitoring & \cite{ren_tinyol_2021} & 
    Collected \\
& & Driver behavior & \cite{silva_adaptive_2023} & 
    Collected \\
\cmidrule{2-5}
& \Block{1-1}{Presence Detection}
  & Multi-sensor: presence detection & 
    \cite{renzi_online_2025} & Collected \\
\cmidrule{2-5}
& \Block{1-1}{Gesture Classification}
  & IMU: on-sensor classification & 
    \cite{chowdhary_-sensor_2023} & Collected \\
\midrule
% ── Concept Drift ─────────────────────────────────────────────
\Block{2-1}{Concept Drift}
  & \Block{2-1}{Sensor Calibration}
  & IMU: calibration & \cite{pau_learning_2025} & 
    Collected, EuRoC MAV \\
& & Pressure sensor: calibration & 
    \cite{saccani_-sensor_2024} & Collected \\
\midrule
% ── Continual Learning ────────────────────────────────────────
\Block{2-1}{Continual Learning}
  & \Block{1-1}{Biosignal Classification}
  & EEG: BMI personalization & \cite{mei_train--request_2024, 
    mei_ultra-low_2025} & Collected \\
\cmidrule{2-5}
& \Block{1-1}{Human Activity Recognition}
  & IMU: class-incremental & \cite{leite_resource-efficient_2022, 
    kwon_exploring_2021} & Opportunity, PAMAP2~\cite{Reiss2012PAMAP2}, Skoda, 
    MHEALTH, Collected \\
\bottomrule
\end{NiceTabular}%
}
\end{table*}

%% file: tables/hardware_ultra_agg.tex
% Table: hardware ultra-constrained (tab:hardware_ultra_agg) — paste content here
\begin{table*}[t]
\centering
\caption{Ultra-Constrained hardware platforms across
surveyed ODL works. $\frequency{}$ and $\memorydevice{}$ report the lower
end of each per-work value. Active power figures are
estimated from device datasheets; not directly reported
in any surveyed work in this class. NR = Not Reported.}
\label{tab:hardware_ultra_agg}
\resizebox{\textwidth}{!}{%
\begin{NiceTabular}{lllp{0.18\textwidth}p{0.10\textwidth}p{0.10\textwidth}p{0.14\textwidth}}[
  code-before =
    \rectanglecolor{envcolor}{2-1}{4-1}
    \rectanglecolor{envcolor}{2-2}{4-2}
    \rectanglecolor{envcolor}{5-1}{6-1}
    \rectanglecolor{envcolor}{5-2}{6-2}
]
\toprule
\textbf{Regime} & \textbf{Origin} & \textbf{Core} &
\textbf{Works} & $\frequency{}$ & $\memorydevice{}$ &
\textbf{Active Power} \\
\midrule
% ── Single Change ─────────────────────────────────────────────
\Block{3-1}{Single-Change}
  & Method
  & Cortex-M0
  & \cite{deutel_-device_2025,
    sudharsan_edge2train_2020,
    sudharsan_ml-mcu_2022,
    sudharsan_train_2021}
  & 48--240~MHz
  & 20--520~kB
  & 10--50~mW \\
\cmidrule{2-7}
  & \Block{2-1}{Application}
  & ISPU
  & \cite{chowdhary_-sensor_2023}
  & 5--10~MHz
  & 8~kB
  & $\sim$1~mW \\
& & Cortex-M0
  & \cite{pereira_-device_2023}
  & 133~MHz
  & 264~kB
  & 20--50~mW \\
\midrule
% ── Concept Drift ─────────────────────────────────────────────
\Block{2-1}{Concept Drift}
  & Method
  & --
  & --
  & --
  & --
  & -- \\
\cmidrule{2-7}
  & Application
  & ISPU
  & \cite{pau_learning_2025}
  & 5--10~MHz
  & 8~kB
  & $\sim$1~mW \\
\bottomrule
\end{NiceTabular}%
}
\end{table*}

%% file: tables/hardware_mcu.tex
% Table: hardware MCU (tab:hardware_mcu) — paste content here
\begin{table*}[t]
\centering
\caption{Standard MCU hardware platforms across surveyed
ODL works. $\frequency{}$ and $\memorydevice{}$ report the lower end of each
per-work value. Active power figures are estimated from
datasheet active-mode figures; not directly reported in
any surveyed work in this class. NR = Not Reported.}
\label{tab:hardware_mcu}
\resizebox{\textwidth}{!}{%
\begin{NiceTabular}{lllp{0.15\textwidth}p{0.13\textwidth}p{0.09\textwidth}p{0.18\textwidth}}[
  code-before =
    \rectanglecolor{envcolor}{2-1}{5-1}
    \rectanglecolor{envcolor}{2-2}{5-2}
    \rectanglecolor{envcolor}{6-1}{7-1}
    \rectanglecolor{envcolor}{6-2}{7-2}
    \rectanglecolor{envcolor}{8-1}{8-1}
    \rectanglecolor{envcolor}{8-2}{8-2}
]
\toprule
\textbf{Regime} & \textbf{Origin} & \textbf{Core} &
\textbf{Works} & $\frequency{}$ & $\memorydevice{}$ &
\textbf{Active Power} \\
\midrule
% ── Single Change ─────────────────────────────────────────────
\Block{4-1}{Single-Change}
  & \Block{2-1}{Method}
  & Cortex-M4
  & \cite{pau_tinyrce_2023, ren_-device_2024,
      khan_dacapo_2023, patil_poet_2022}
  & 180~MHz
  & 250~kB
  & 20--300~mW \\
& & Cortex-M7
  & \cite{lin_-device_2024, pau_tinyrce_2023-1,
      disabato_incremental_2020}
  & 160--216~MHz
  & 64--512~kB
  & 100--500~mW \\
\cmidrule{2-7}
  & \Block{2-1}{Application}
  & Cortex-M4
  & \cite{pavan_tinysv_2025, craighero_-device_2024,
      abdennadher_fixed_2021, ren_tinyol_2021}
  & 64--150~MHz
  & 60--400~kB
  & 20--60~mW \\
& & Cortex-M7
  & \cite{pau_online_2021}
  & NR
  & NR
  & NR \\
\midrule
% ── Concept Drift ─────────────────────────────────────────────
\Block{2-1}{Concept Drift}
  & \Block{2-1}{Method}
  & Cortex-M4
  & \cite{pavan_tybox_2024}
  & 64~MHz
  & 256~kB
  & 20--50~mW \\
& & Cortex-M7
  & \cite{disabato_tiny_2024}
  & 84~MHz
  & 96~kB
  & 300--500~mW \\
\midrule
% ── Continual Learning ────────────────────────────────────────
\Block{1-1}{Continual Learning}
  & Method
  & Cortex-M7
  & \cite{kwon_lifelearner_2023}
  & 480~MHz
  & 200~kB
  & 250--600~mW$^{a}$ \\
\bottomrule
\end{NiceTabular}%
}
\par\smallskip\footnotesize $^{a}$The upper end (600~mW) marginally exceeds the 500~mW criterion in Section~\ref{sect:inclusion}. The device (STM32H7 at 480~MHz) is MCU-class in all other respects; 600~mW is a worst-case datasheet estimate across supply-voltage configurations, not a measured workload power. This work is included on that basis.
\end{table*}

%% file: tables/hardware_pulp.tex
% Table: hardware PULP (tab:hardware_pulp) — paste content here

\begin{table*}[t]
\centering
\caption{PULP hardware platforms across surveyed ODL works.
$\frequency{}$ and $\memorydevice{}$ report the lower end of each per-work value.
Active power figures are reported from measurements where
available. NR = Not Reported.}
\label{tab:hardware_pulp}
\resizebox{\textwidth}{!}{%
\begin{NiceTabular}{lllp{0.12\textwidth}p{0.13\textwidth}p{0.13\textwidth}p{0.20\textwidth}}[
  code-before =
    \rectanglecolor{envcolor}{2-1}{5-1}
    \rectanglecolor{envcolor}{2-2}{5-2}
    \rectanglecolor{envcolor}{6-1}{8-1}
    \rectanglecolor{envcolor}{6-2}{8-2}
]
\toprule
\textbf{Regime} & \textbf{Origin} & \textbf{Device} &
\textbf{Works} & $\frequency{}$ & $\memorydevice{}$ &
\textbf{Active Power} \\
\midrule
% ── Single Change ─────────────────────────────────────────────
\Block{4-1}{Single-Change}
  & \Block{4-1}{Application}
  & GAP8
  & \cite{burrello_tackling_2021}
  & 100~MHz
  & 32~MB
  & NR \\
& & GAP9
  & \cite{cereda_training_2024, cereda_-device_2024,
      cioflan_-device_2024, rusci_self-learning_2025,
      rusci_-device_2023, zanghieri_semg-driven_2024,
      rusci_self-incremental_2025}
  & 240--370~MHz
  & 128~kB--10~MB
  & 19--66~mW \\
& & VEGA
  & \cite{cioflan_boosting_2024, cioflan_towards_2022}
  & NR
  & 15~kB--1.5~MB
  & NR \\
& & Mr.~Wolf
  & \cite{benatti_online_2019}
  & 400~MHz
  & $<$1~MB
  & 10.4~mW \\
\midrule
% ── Continual Learning ────────────────────────────────────────
\Block{3-1}{Continual Learning}
  & \Block{2-1}{Method}
  & GAP9
  & \cite{wibowo_12_2024}
  & 240~MHz
  & 10~MB
  & 45~mW \\
& & VEGA
  & \cite{ravaglia_tinyml_2021}
  & 375~MHz
  & 64~MB
  & NR \\
\cmidrule{2-7}
  & Application
  & GAP9
  & \cite{mei_train--request_2024, mei_ultra-low_2025}
  & 370~MHz
  & 1--4~MB
  & 21--50~mW \\
\bottomrule
\end{NiceTabular}%
}
\end{table*}

%% file: tables/batch_learning.tex
% Table: batch learning (tab:batch_learning) — paste content here
\begin{table*}[t]
\centering
\caption{Batch learning ODL works, organized by 
origin and device class.}
\label{tab:batch_learning}
\resizebox{\textwidth}{!}{%
\begin{NiceTabular}{p{0.07\textwidth}lp{0.09\textwidth}p{0.08\textwidth}p{0.11\textwidth}p{0.18\textwidth}p{0.14\textwidth}p{0.09\textwidth}}[
  code-before =
    \rectanglecolor{envcolor}{2-1}{20-3}
    \rectanglecolor{envcolor}{2-2}{9-2}
    \rectanglecolor{envcolor}{2-3}{9-3}
]
\toprule
\textbf{Regime} & \textbf{Origin} & \textbf{Device} & 
\textbf{Works} & \textbf{$\feblock{}$} & 
\textbf{$\filterblock{}$} & \textbf{$\learningmech{}$} & 
\textbf{$\algo{}$} \\
\midrule
% ── Single Change, Method-oriented ───────────────
\Block{17-1}{Single-Change}
  & \Block{9-1}{Method}
  & \Block{3-1}{Ultra-Constr.}
    & \cite{deutel_-device_2025}              
    & --- & All data in mem. 
    & Quant. backprop & Q-DNN \\
& & & \cite{sudharsan_edge2train_2020}         
    & --- & All data in mem. 
    & SVM train alg. & SVM \\
& & & \cite{sudharsan_ml-mcu_2022}             
    & --- & All data in mem. 
    & SGD / 1-v-1 & Linear cls. \\
\cmidrule{3-8}
  & & \Block{2-1}{Std.\ MCU}
    & \cite{khan_dacapo_2023}                  
    & CNN & All data in mem. 
    & Backprop & DNN \\
& & & \cite{lin_-device_2024}                  
    & CNN & All data in mem. 
    & Sparse backprop & DNN \\
\cmidrule{3-8}
  & & \Block{4-1}{No device}
    & \cite{cai_tinytl_2021}                   
    & --- & All data in mem. 
    & Bias-only backprop & DNN \\
& & & \cite{rub_tinyprop_2023, rub_advancing_2024} 
    & --- & All data in mem. 
    & Sparse backprop & DNN \\
& & & \cite{kwon_tinytrain_2024}               
    & CNN & All data in mem. 
    & Sparse backprop & DNN \\
& & & \cite{rub_sparse_2025}                   
    & --- & All data in mem. 
    & Sparse fwd-fwd & DNN \\
\cmidrule{2-8}
% ── Single Change, Application-oriented ──────────
  & \Block{8-1}{Appl.}
  & \Block{1-1}{Ultra-Constr.}
    & \cite{pereira_-device_2023}              
    & --- & All data in mem. 
    & EM param.\ fit & Weibull \\
\cmidrule{3-8}
  & & \Block{3-1}{Std.\ MCU}
    & \cite{abdennadher_fixed_2021}            
    & Reservoir+PCA & All data in mem. 
    & SVM training (SMO) & 1-class SVM \\
& & & \cite{craighero_-device_2024}            
    & CNN & All data in mem. 
    & Backprop & DNN \\
& & & \cite{pavan_tinysv_2025}                 
    & MFCC+CNN & Few labeled ex. 
    & Feature storage & KNN-like \\
\cmidrule{3-8}
  & & \Block{4-1}{PULP}
    & \cite{cereda_training_2024, cereda_-device_2024} 
    & CNN & All data in mem. 
    & Backprop & MLP \\
& & & \cite{cioflan_boosting_2024, cioflan_-device_2024, 
      cioflan_towards_2022} 
    & MFCC+CNN & New + old data 
    & Backprop & DNN \\
& & & \cite{rusci_-device_2023}                
    & MFCC+CNN & Few ex.\ per class 
    & Prototype gen. & Prototype \\
& & & \cite{rusci_self-learning_2025}          
    & MFCC+CNN & Balanced pos./neg. 
    & Backprop + prototypes gen. & Prototype \\
\cmidrule{3-8}
  & & \Block{1-1}{No device}
    & \cite{rusci_few-shot_2023}               
    & MFCC+CNN & Few ex.\ per class 
    & Prototype gen. & Prototype \\
\bottomrule
\end{NiceTabular}%
}
\end{table*}

%% file: tables/incremental_learning.tex
% Table: incremental learning (tab:incremental_learning) — paste content here
\begin{table*}[t]
\centering
\caption{Incremental Learning ODL works, organized by environment, origin, and device type.
\textbf{SC} = Single-Change, \textbf{CD} = Concept Drift, \textbf{CL} = Continual Learning.}
\label{tab:incremental_learning}
\resizebox{\textwidth}{!}{%
\begin{NiceTabular}{p{0.07\textwidth}lp{0.11\textwidth}p{0.07\textwidth}p{0.11\textwidth}p{0.18\textwidth}p{0.14\textwidth}p{0.09\textwidth}}[
  code-before =
    \rectanglecolor{envcolor}{2-1}{25-3}
]
\toprule
\textbf{Reg.} & \textbf{Origin} & \textbf{Device} & \textbf{Works} & \textbf{R} & \textbf{B} & \textbf{L} & \textbf{f} \\
\midrule
% ── Single Change ──────────────────────────────────────────────
\Block{10-1}{{Single-Change}}
  & \Block{5-1}{Method}
  & Ultra-Const.     & \cite{sudharsan_train_2021}                          & ---      & Online (|$\Delta$|=1)         & SGD (binary cls.)              & Linear cls. \\
\cmidrule{3-8}
  & & \Block{3-1}{Std.\ MCU}
    & \cite{disabato_incremental_2020}                     & CNN      & Online (|$\Delta$|=1)         & Feat. vec. store         & KNN \\
& & & \cite{pau_tinyrce_2023, pau_tinyrce_2023-1}          & CNN      & Fixed buf., emptied on trigger & Fwd-only (RCE update)         & RCE cls. \\
& & & \cite{ren_-device_2024}                              & CNN      & Online (|$\Delta$|=1)         & Backprop                       & DNN \\
\cmidrule{3-8}
  & & No device
    & \cite{pavan_tactile_2025, rub_drip_2025}             & CNN      & Active learning buf.\ selection & Backprop                     & DNN \\
\cmidrule{2-8}
  & \Block{5-1}{Appl.}
  & Ultra-Const.     & \cite{chowdhary_-sensor_2023}                        & Stat.\ FE & Online (|$\Delta$|=1)        & Prototype gen.         & Prototype \\
\cmidrule{3-8}
  & & Std.\ MCU         & \cite{ren_tinyol_2021}                               & CNN      & Online (|$\Delta$|=1)         & Backprop                       & DNN \\
\cmidrule{3-8}
  & & \Block{2-1}{PULP}
    & \cite{burrello_tackling_2021}                        & TCN      & Fixed buf.\ (uniform + exp. sampling)  & Backprop                       & TCN \\
& & & \cite{zanghieri_semg-driven_2024}                    & ---      & Online (|$\Delta$|=1)         & Backprop                       & TCN \\
\cmidrule{3-8}
  & & No device         & \cite{renzi_online_2025}                             & ---      & Online (|$\Delta$|=1)         & Various       & Various \\
\midrule
% ── Concept Drift ──────────────────────────────────────────────
\Block{6-1}{{Concept Drift}}
  & \Block{4-1}{Method}
  & \Block{2-1}{Std.\ MCU}
    & \cite{disabato_tiny_2024}                            & CNN      & Fixed buf. + change detect. and removal & Feature vec. storage\  & KNN-like \\
& & & \cite{pavan_tybox_2024}                              & CNN+Stat.\ FE & Fixed buf.                & Backprop                       & DNN \\
\cmidrule{3-8}
  & & \Block{2-1}{No device}
    & \cite{ma_-demand_2025}                               & ---      & Fixed buf.\ + change detect.  & BN adapt.\ + model select.     & Multi-CNN \\
& & & \cite{lourenco_dfdt_2025}                            & ---      & Online (|$\Delta$|=1)         & Hoeffding-style tree update    & Decision tree \\
\cmidrule{2-8}
  & \Block{1-1}{Appl.}
  & Ultra-Const.     & \cite{pau_learning_2025, saccani_-sensor_2024}                             & ---      & Online (|$\Delta$|=1)         & Fwd-only upd.          & RBF-NN \\
\midrule
% ── Continual Learning ─────────────────────────────────────────
\Block{8-1}{{Continual Learning}}
  & \Block{5-1}{Method} & Std.\ MCU         & \cite{kwon_lifelearner_2023}                         & CNN (meta) & Fixed buf.\ per class       & Backprop                       & DNN \\
\cmidrule{3-8}
  & & \Block{2-1}{PULP}
    & \cite{ravaglia_tinyml_2021}                          & CNN      & Fixed buf. & Backprop  & DNN \\
& & & \cite{wibowo_12_2024}                                & CNN (meta) & Fixed buf.\ per class       & Prototype gen.         & Prototype \\
\cmidrule{3-8}
  & & \Block{2-1}{No device}
    & \cite{vorabbi_enabling_2024}                         & CNN      & Fixed buf.\ (binary)  & Backprop       & BNN \\
& & & \cite{rub_continual_2024}                            & ---      & Fixed buf.\ (distilled)  & Backprop                       & DNN \\
\cmidrule{2-8}
  & \Block{3-1}{Appl.} & \Block{2-1}{PULP}
    & \cite{mei_train--request_2024}                       & Filtering & Fixed buf., trigger on acc.\ drop & Backprop                  & CNN \\
& & & \cite{mei_ultra-low_2025}                            & CNN      & Fixed buf.                    & Backprop                       & DNN \\
\cmidrule{3-8}
  & & No device         & \cite{leite_resource-efficient_2022}                 & ---      & Fixed buf.                    & Backprop + net.\ expansion     & Expandable DNN \\
\bottomrule
\end{NiceTabular}%
}
\end{table*}

%% file: tables/components.tex
% Table: components (tab:components) — paste content here
\begin{table*}[t]
\centering
\caption{Component choices for $\solution{}$, with
qualitative $\memory{}_{\solution{}}^{\ti{}}$ and $\executiontime{}(\solution{}^{\ti{}}, \device{})$ for each component. Categories
are relative within each component row: \textbf{L} =
low, \textbf{M} = medium, \textbf{H} = high.
$\uparrow$ indicates that the metrics grows
over time as new data are incorporated. $\executiontime{}(\solution{}^{\ti{}}, \device{})$
refers to inference cost for $\feblock{}$ and
$\algo{}$, and to learning cost for $\learningmech{}$.
$\filterblock{}$ does not contribute to $\executiontime{}(\solution{}^{\ti{}}, \device{})$
directly.}
\label{tab:components}
\resizebox{0.6\textwidth}{!}{%
\begin{NiceTabular}{llcc}
\toprule
\textbf{Component} & \textbf{Variant} &
$\memory{}_{\solution{}}^{\ti{}}$ & $\executiontime{}(\solution{}^{\ti{}}, \device{})$ \\
\midrule
% ── Dimensionality Reduction Block ───────────────
\Block{3-1}{$\feblock{}$}
  & Identity ($\feblock{} = \text{id}$) 
  & L & L \\
& Statistical FE 
  & L & M \\
& Frozen pre-trained CNN 
  & H & M-H \\
\midrule
% ── Data Management Block ────────────────────────
\Block{3-1}{$\filterblock{}$}
  & Online ($|\filterbatch{}| = 1$) 
  & L & -- \\
& Explicit buffer ($|\filterbatch{}| > 1$) 
  & M & -- \\
& Full batch collection 
  & H & -- \\
\midrule
% ── ML Algorithm ─────────────────────────────────
\Block{5-1}{$\algo{}$}
  & Prototype 
  & L$\uparrow$ & L$\uparrow$ \\
& KNN 
  & M$\uparrow$ & M-H$\uparrow$ \\
& DNN 
  & M--H & M \\
& CNN 
  & H & H \\
& TNN 
  & H & H \\
\midrule
% ── Learning Mechanism ───────────────────────────
\Block{6-1}{$\learningmech{}$}
  & Prototype construction 
  & L & L \\
& Add to KNN pool
  & L & L \\
& Forward-only update 
  & L & M \\
& Full backpropagation 
  & H & H \\
& Sparse backpropagation 
  & M & M \\
& Bias-only / head-only update 
  & L & M \\
\bottomrule
\end{NiceTabular}%
}
\end{table*}

%% file: tables/reporting.tex
% Table: reporting (tab:reporting) — paste content here
\begin{table*}[t]
\centering
\caption{Normalization conventions used for latency 
and energy reporting across the surveyed works, 
organized by distribution change regime. 
SC~=~Single-change, CD~=~Concept Drift, 
CL~=~Continual Learning. }
\label{tab:reporting}
\resizebox{\textwidth}{!}{%
\begin{NiceTabular}{llp{0.28\textwidth}p{0.14\textwidth}p{0.18\textwidth}}
\toprule
\textbf{Metric} & \textbf{Unit} & \textbf{SC} & 
\textbf{CD} & \textbf{CL} \\
\midrule
\Block{5-1}{Latency}
  & Per instance
  & \cite{zanghieri_semg-driven_2024, ren_tinyol_2021,
      disabato_incremental_2020, ren_-device_2024,
      deutel_-device_2025, chowdhary_-sensor_2023}
  & \cite{disabato_tiny_2024}
  & \cite{mei_ultra-low_2025} \\
& Per batch
  & \cite{craighero_-device_2024}
  & \cite{pavan_tybox_2024}
  & -- \\
& Per epoch
  & \cite{cereda_training_2024, cereda_-device_2024,
      lin_-device_2024}
  & --
  & -- \\
& Per learning event
  & --
  & --
  & \cite{ravaglia_tinyml_2021, kwon_lifelearner_2023} \\
& Per complete run
  & \cite{cioflan_-device_2024, cioflan_towards_2022,
      rusci_self-learning_2025, rusci_-device_2023,
      benatti_online_2019, burrello_tackling_2021,
      abdennadher_fixed_2021, pavan_tinysv_2025,
      khan_dacapo_2023, sudharsan_edge2train_2020,
      sudharsan_ml-mcu_2022, sudharsan_train_2021}
  & \cite{pau_learning_2025}
  & \cite{mei_train--request_2024, wibowo_12_2024} \\
\midrule
\Block{5-1}{Energy}
  & Per instance
  & \cite{zanghieri_semg-driven_2024, ren_-device_2024}
  & --
  & -- \\
& Per batch
  & \cite{craighero_-device_2024, khan_dacapo_2023}
  & --
  & -- \\
& Per epoch
  & \cite{cereda_training_2024, cioflan_boosting_2024}
  & --
  & -- \\
& Per learning event
  & --
  & --
  & \cite{ravaglia_tinyml_2021, kwon_lifelearner_2023} \\
& Per complete run
  & \cite{cioflan_-device_2024, cioflan_towards_2022,
      rusci_self-learning_2025, rusci_-device_2023,
      benatti_online_2019, burrello_tackling_2021,
      sudharsan_edge2train_2020, sudharsan_train_2021}
  & --
  & \cite{wibowo_12_2024} \\
\bottomrule
\end{NiceTabular}%
}
\end{table*}

%% file: tables/notation.tex
% Table: notation (tab:notation) — paste content here
\begin{table}[h]
\centering
\caption{Mathematical notation used throughout the paper.}
\label{tab:notation}
\begin{tabular}{lp{0.6\columnwidth}}
\toprule
\textbf{Symbol} & \textbf{Meaning} \\
\midrule
\multicolumn{2}{l}{\textit{Problem formulation}} \\
$\dataprocess{}$ & Data-generating process \\
$\device{}$ & Target device \\
$\solution{}$ & ODL solution \\
$\deploytime{}$ & Deployment time \\
$\trainset{}$ & Pre-deployment dataset \\
$\testset{}$ & Held-out test set \\
$\stream{}$ & Post-deployment data stream \\
\midrule
\multicolumn{2}{l}{\textit{Device and application constraints}} \\
$\memorydevice{}$ & Device RAM capacity \\
$\memory{}_{\solution{}}^{t}$ & RAM used by $\solution{}$ at time $t$ \\
$\frequency{}$ & Device clock frequency \\
$\executiontime{}_{\solution{}}^{t}$ & Execution time of $\solution{}$ at time $t$ \\
$\executionconstraint{}$ & Latency constraint \\
\midrule
\multicolumn{2}{l}{\textit{Solution components}} \\
$\feblock{}$ & Dimensionality reduction block \\
$\latentrep{}^t$ & Latent representation at time $t$ \\
$\filterblock{}$ & Data management block \\
$\filterbatch{}^t$ & Data management buffer at time $t$ \\
$\algo{}^t$ & ML algorithm at time $t$ \\
$\learningmech{}$ & Learning mechanism \\
\midrule
\multicolumn{2}{l}{\textit{Data and evaluation}} \\
$\inp{}$ & Input observation at time $t$ \\
$\pred{}$ & Prediction at time $t$ \\
$\metric{}$ & Performance metric \\
$\adaptsetdim{}$ & Adaptation set size \\
\bottomrule
\end{tabular}
\end{table}